%% file: main.tex
\documentclass{article}
\usepackage[T1]{fontenc}
\usepackage[utf8]{inputenc}
\usepackage[margin=1in]{geometry}
\usepackage{amsmath,amssymb,amsthm,mathtools}
\usepackage{graphicx}
\usepackage{subcaption}
\usepackage{float}
\usepackage{enumitem}
\usepackage[square,authoryear]{natbib}
\usepackage{xcolor}
\usepackage[colorlinks=true,linkcolor=blue,citecolor=blue,urlcolor=blue]{hyperref}

\input{math_commands}
\input{defs}

\newcommand{\norm}[1]{\left\|#1\right\|}
\newcommand{\ip}[2]{\left\langle #1,#2\right\rangle}

\newtheorem{theorem}{Theorem}
\newtheorem{lemma}{Lemma}
\newtheorem{proposition}{Proposition}

\usepackage{xcolor}
\definecolor{DarkGreen}{rgb}{0.1,0.5,0.1}
\definecolor{DarkRed}{rgb}{0.5,0.1,0.1}
\definecolor{DarkBlue}{rgb}{0.1,0.1,0.5}
\definecolor{DarkYellow}{rgb}{.79,.79,0}
\definecolor{unitednationsblue}{rgb}{0.36, 0.57, 0.9}
\definecolor{blue_ppt}{rgb}{0,0.6,0.93}
\definecolor{darkblue_ppt}{rgb}{0.05,0.4,0.8}
\definecolor{orange_ppt}{rgb}{0.82,0.5,0}
\usepackage{color}
\definecolor{violet}{RGB}{138,43,226}

\renewcommand{\paragraph}[1]{\par\medskip\noindent\textbf{#1}\quad\ignorespaces}

\title{Stochastic Inertial Krasnosel'skii--Mann Iteration\\
Achieves Near-Optimal Sample Complexity}
\author{%
 Tong Yang\thanks{Department of Electrical and Computer Engineering, Carnegie Mellon University; email: \texttt{tongyang@andrew.cmu.edu}. }\textsuperscript{\phantom{\textasteriskcentered},}\thanks{Fundamental AI Research (FAIR), Meta Superintelligence Labs; emails: \texttt{\{taojiang,linx\}@meta.com}. } \\
 CMU \& Meta FAIR \\
 \and
 Tao Jiang\footnotemark[2] \\
 Meta FAIR \\
 \and
 Yuejie Chi\thanks{Department of Statistics and Data Science, Yale University; email: \texttt{yuejie.chi@yale.edu}. } \\
 Yale University\\
 \and
 Ashok Cutkosky\thanks{Fundamental AI Research (FAIR), Meta Superintelligence Labs; Department of Electrical and Computer Engineering, Boston University; email: \texttt{ashok@cutkosky.com}. } \\
  Meta FAIR \& Boston University \\
 \and
 Lin Xiao\footnotemark[2] \\
 Meta FAIR \\
}
\date{\today}

\begin{document}
\maketitle

\begin{abstract}
We analyze a simple stochastic inertial Krasnosel'skii--Mann (iKM) method
for finding a fixed point of a nonexpansive operator in a real Hilbert space.
Our method is obtained simply by adding two inertial extrapolations to
stochastic KM~\citep{bravo2024stochastic}, and it retains one call to a possibly
biased stochastic oracle per update and achieves sharp rates in both the
stochastic and deterministic regimes. Specifically, with our proposed parameter schedule,
we prove the following last-iterate fixed-point residual bound:
\[
 {O}\!\left(\frac{1}{K}
+\frac{\sigma\log K}{\sqrt K}
+\frac{B_K\log K}{K}\right),
\]
where $K$ is the horizon, $\sigma$ is the noise level and $B_K$ is the accumulated
root-mean-square bias. When $B_K=O(\sqrt K)$, this yields
$\widetilde O(\epsilon^{-2})$ sample complexity that
matches, up to a logarithmic factor, the stochastic-oracle lower bound given under the unbiased subclass of our model~\citep[Theorem~2]{foster2019complexity}. It also improves the best-known $O(\epsilon^{-4})$ random-iterate guarantee for stochastic KM~\citep[Corollary~5.4]{bravo2024stochastic}. To our knowledge, this is the
first single-loop method for general nonexpansive fixed-point problems to
attain this near-optimal sample complexity without variance reduction or
batching.
When the oracle is exact, the
same method attains the worst-case-optimal $O(K^{-1})$ last-iterate residual rate
\citep[Theorem~4.6]{park2022exact}, improving the $O(K^{-1/2})$
rate of classical KM
\citep{cominetti2014rate,bravo2018sharp}.
\end{abstract}

\section{Introduction}
\label{sec:introduction}

We consider a stochastic fixed-point problem for a nonexpansive operator
$T:\gH\to\gH$ in a real Hilbert space.  Assuming that its fixed-point set
$\Fix T\coloneqq\{x\in\gH:Tx=x\}$ is nonempty, the goal
is to find an approximate solution of
\[
 Tp^\star=p^\star,
\]
with accuracy measured by the fixed-point residual. Such nonexpansive fixed-point equations provide a common formulation for convex
feasibility, optimization, and equilibrium problems expressed as monotone
inclusions~\citep{bauschke2017convex,cortild2026stochastic}. For example, in optimization,
$T$ can be the update map of a first-order or operator-splitting method, so
its fixed points can encode solutions, zeros, or stationary points of the
underlying problem. The fixed-point residual -- which is used to measure the final accuracy -- then specializes to a familiar feasibility, stationarity, or equilibrium measure in these applications.

In many modern applications dealing with large-scale or expectation-based settings, the exact evaluation of applying the operator
$T$ may be unavailable or
prohibitively costly, so sampling and approximate computation naturally
produce noisy and potentially biased responses
\citep{bravo2024stochastic,cui2019inexact}.
Consequently, we focus on the stochastic case where the operator $T$ is not available exactly: at each query, the algorithm receives a
noisy and possibly biased response. Given a target precision $\epsilon>0$, we study how many such oracle responses are needed to make the {\em last-iterate} residual at most $\epsilon$. 

Perhaps the most common approach to
stochastic fixed-point iterations are obtained by replacing the
exact operator evaluations in classical deterministic schemes with
stochastic oracle responses.  These include the Krasnosel'skii--Mann (KM)
iteration~\citep{mann1953mean,krasnoselskii1955two} with update
\begin{equation}
 x_{k+1}=(1-\lambda_k)x_k+\lambda_kTx_k,
 \label{eq:km_iteration}
\end{equation}
the anchored iteration of
\citet{halpern1967fixed} with update
\begin{equation}
 x_{k+1}=(1-\lambda_k)x_0+\lambda_kTx_k,
 \label{eq:halpern_iteration}
\end{equation} 
and their accelerated variants
\citep{mainge2008convergence,dong2018general,bot2023fast,park2022exact}. Some of these schemes attain the optimal rate
\citep{contreras2022optimal,park2022exact} under the deterministic setting, see Section~\ref{sec:related-work} for details.  However, under stochastic evaluations, existing results for general nonexpansive maps reveal a
simplicity--complexity gap: simple methods have substantially suboptimal sample complexity, whereas near-optimal guarantees require stronger assumptions or
more sophisticated algorithms. We next elaborate on this trade-off more quantitatively, as follows. To begin, let us record a useful lower bound for the stochastic
fixed-point problem studied here: given a target precision $\epsilon$, \citet[Theorem~2]{foster2019complexity} established that at least $\Omega(\epsilon^{-2})$ samples are needed, regardless of algorithms in use.
\begin{itemize}
\item \textbf{Substantially sub-optimal rates for simple single-loop algorithms.} For stochastic KM iteration,
\citet[Theorem~4.1 and Corollary~5.4]{bravo2024stochastic} obtained
sample-complexity bounds of $O(\epsilon^{-6})$ and $O(\epsilon^{-4})$ for
the expected residual of the last iterate and a weighted random iterate,
respectively.  \citet[Corollary~2.10]{cortild2026stochastic} also obtained the
$O(\epsilon^{-4})$ rate for one-sample stochastic KM.
For stochastic Halpern iteration,
\citet[Corollary~3.5]{bravo2026stochastic} and
\citet[Section~5 and Example~5.1]{pischke2026asymptotic} gave last-iterate
sample-complexity bounds of $\widetilde O(\epsilon^{-5})$, using minibatches
of size $\widetilde O(\epsilon^{-4})$. Under cocoercivity in expectation,
S-Dual-OHM~\citep{yoon2026direct} achieved a sample complexity of $O(\epsilon^{-3})$ using minibatches of size $O(\epsilon^{-2})$. None of these methods achieve the optimal order $\widetilde O(\epsilon^{-2})$ of sample complexity.

\item \textbf{Sophisticated algorithm designs to accelerate convergence.}  For general fixed-point
equations, VR-GHAL~\citep[Algorithm~3 and Corollary~3]{diakonikolas2026solving} reached
$\widetilde O(\epsilon^{-2})$ complexity under samplewise nonexpansiveness,
but required an array of techniques including recursive variance reduction, same-seed difference queries, nested epochs,
and large refresh batches. 
\citet[Section~4.2 and Corollary~2]{nguyentrung2025class} used growing minibatches and has suboptimal sample complexity.  
% In finite-dimensional stochastic monotone problems, RAIN
% \citep[Theorem~4.2]{chen2024near} attains near-optimal complexity using a
% nested, recursively anchored regularization scheme built from epochwise
% stochastic extragradient.
Under
related cocoercive assumptions, the recursively variance-reduced Halpern methods of
\citet{cai2022stochastic} attained $O(\epsilon^{-3})$ complexity in the general regime. 
% Hence near-optimal stochastic dependence is attainable, but the available
% guarantees obtain it through additional oracle structure, batching, variance
% reduction, nested procedures, or stronger problem assumptions.
\end{itemize}

This simplicity--complexity gap motivates the following question:

\begin{center}
\begin{minipage}{0.88\linewidth}
\centering\itshape
Can a simple single-loop method attain near-optimal sample complexity for
general nonexpansive stochastic fixed-point problems?
\end{minipage}
\end{center}

Inspired by the general inertial Krasnosel'skii--Mann (iKM) iteration introduced by
\citet[Section~3.2, Eq.~(18), p.~181]{dong2018general}, we study its
one-sample stochastic counterpart, and with carefully chosen parameter schedule,  provide a positive answer to the above question. 

\subsection{Our contributions}
We analyze a simple stochastic inertial Krasnosel'skii--Mann (iKM) method, which
adds two inertial extrapolations to
stochastic KM as follows:
\begin{equation*} 
\begin{aligned}
 y_k&=x_k+\alpha_k(x_k-x_{k-1}),\\
 z_k&=x_k+\beta_k(x_k-x_{k-1}),\\
 x_{k+1}&=(1-\lambda_k)y_k+\lambda_k\widehat T_k(z_k),
 \qquad k=1,\ldots,K.
\end{aligned}
\end{equation*}
Here, \(\alpha_k\geq 0 \) and \(\beta_k \geq 0 \) are the two inertial
extrapolation coefficients associated with \(y_k\) and \(z_k\), respectively, \(\lambda_k \in (0,1] \) is the relaxation parameter, and $\widehat T_k$ is a noisy, and possibly biased, evaluation of the fixed-point operator $T$. This algorithm is simple with a single-loop structure, making it highly implementable in practice. In particular, each update retains exactly one stochastic oracle call, consistent with stochastic KM.

With carefully designed schedule, 
we establish simultaneous fixed-point-residual bounds for the last
 iterate $x_K$ and the two corresponding inertial extrapolations $y_K, z_K$ of iKM on the order of
\[
 O\!\left(\frac{1}{K}
+\frac{\sigma\log K}{\sqrt K}
+\frac{B_K\log K}{K}\right),
\]
where $K$ is the horizon, $\sigma$ is the noise level and $B_K$ is the accumulated
root-mean-square bias.
\begin{itemize}%[leftmargin=*,itemsep=0.8em]
 %\item \textbf{A simple stochastic iKM method.} We consider a standard stochastic oracle model where the oracle may be biased and has zero-mean, bounded-variance noise. The stochastic iKM method we propose is simple, single-loop and highly implementable:relative to stochastic KM, each update adds only two affine inertial extrapolations, while retaining exactly one stochastic oracle call.

 \item \textbf{Near-optimal stochastic sample complexity.}
When the accumulated
bias $B_K$ is controlled by $O(\sqrt{K})$, iKM has $\widetilde O(\epsilon^{-2})$ sample complexity, matching the lower bound in
 \citep[Theorem~2]{foster2019complexity} for unbiased oracles up to one logarithmic factor, and significantly improving the best-known rate $O(\epsilon^{-4})$ 
  of stochastic KM, which holds for a weaker random-iterate convergence %under a related oracle model
 \citep[Corollary~5.4]{bravo2024stochastic}.  

 \item \textbf{Optimal deterministic specialization.}
 In the special case where the oracle is exact, namely $\widehat{T}_k = T$ for all $k$ with $\sigma=0$ and $B_K = 0$, iKM attains a faster
 $O(K^{-1})$ last-iterate rate, which is worst-case optimal for
 deterministic nonexpansive fixed-point methods
 \citep[Theorem~4.6]{park2022exact}, and strictly improves the sharp
 $\Theta(K^{-1/2})$ order of classical KM \citep{davis2016convergence}.  

\end{itemize}

To the best of our knowledge, our work provides the
 first single-loop method for general nonexpansive fixed-point problems to
 attain the near-optimal sample complexity for the stochastic setting, without sophisticated algorithmic operations such as variance reduction or batching. The same algorithm also achieves optimality in the deterministic setting, therefore smoothly interpolating both regimes.

\subsection{Related work}\label{sec:related-work}

\paragraph{Classical deterministic fixed-point methods.}
Under exact evaluations, the classical KM iteration has a sharp
$O(K^{-1/2})$ worst-case last-iterate residual rate for general nonexpansive
operators \citep{cominetti2014rate,bravo2018sharp}.  The anchored Halpern
iteration can attain the faster $O(K^{-1})$ rate, with tight guarantees in
Hilbert and general normed spaces
\citep{lieder2021convergence,contreras2022optimal}; this order is worst-case
optimal in the deterministic fixed-point oracle model
\citep[Theorem~4.6]{park2022exact}.  Related refinements include rates for
operator-splitting schemes \citep{davis2016convergence} and, in finite
dimensions, continuous families of methods with the same exact optimal
guarantee \citep{yoon2024optimal}.  These results assume exact evaluations
and do not quantify the effect of stochastic oracle errors.

\paragraph{Inertial and accelerated KM methods.}
We consider the two-extrapolation general iKM iteration originated from
\citet[Section~3.2, Eq.~(18), p.~181]{dong2018general}, which recovers the earlier
common-extrapolation scheme of \citet{mainge2008convergence} when its two
extrapolation coefficients are equal.  Prior work establishes deterministic
convergence for common- and two-extrapolation variants
\citep{cui2019inexact,maulen2024inertial,cortild2025krasnoselskii}.  Besides, \citet{bot2023fast,bot2026overrelaxation} propose Fast KM - an accelerated version of KM with optimal  $O(K^{-1})$ deterministic last-iterate rate.
Appendix~\ref{app:relation-fast-km} shows that the Fast-KM iteration can be seen as
deterministic iKM under a particular hyperparameter choice.  These
analyses concern exact evaluations or summable perturbations and do not
account for stochastic-oracle complexity.

\paragraph{Stochastic fixed-point methods.}
Existing direct stochastic KM and Halpern guarantees fall short of
near-optimal sample complexity: available one-sample KM bounds are
$O(\epsilon^{-4})$ or worse
\citep{bravo2024stochastic,cortild2026stochastic}, while minibatched analyses
rely on growing batches
\citep{iiduka2026minibatch,bravo2026stochastic,pischke2026asymptotic}.
Near-optimal rates instead require additional structure or machinery.
VR-GHAL~\citep[Algorithm~3 and Corollary~3]{diakonikolas2026solving} reaches
$\widetilde O(\epsilon^{-2})$ under sample-wise nonexpansiveness using
recursive variance reduction, which requires a stronger stochastic oracle supporting same-seed difference queries.
The restarted variance-reduced Halpern method of \citet{cai2022stochastic}
achieves $O(\epsilon^{-2}\log(1/\epsilon))$ only under an additional
sharpness-type error bound, while
variance-reduction-free S-Dual-OHM~\citep{yoon2026direct} reaches
$\widetilde O(\epsilon^{-2})$ only under the stronger assumption that the
residual operator $I-T$ is strongly monotone.  The stochastic
AFP method of \citet[Section~4.2 and Corollary~2]{nguyentrung2025class} is
single-loop but uses polynomially growing minibatches. A particularly relevant comparison is the RAIN algorithm of \citet{chen2024near}, developed for stochastic minimax problems. After translating to our setting, RAIN achieves $O(\log^{3}(1/\epsilon)/\epsilon^2)$ oracle complexity instead of our $O(\log^2(1/\epsilon)/\epsilon^2)$ rate. RAIN employs a recursive regularization scheme similar to SGD3~\citep{allen2018make} (augmented with an extra oracle call per iteration, inheriting both the $\log^3$ term and also a double-loop structure), while we take a different single-loop, single-oracle call path based on iKM. Our method also accommodates general unknown bias, does not require the distance $D=\operatorname{dist}(x_1,\Fix T)$ as an algorithmic input, and obtains a last-iterate guarantee. 
%single-loop but uses polynomially growing minibatches. A particularly relevant comparison is the RAIN algorithm of \citet{chen2024near}, developed for stochastic minimax problems. After translating to our setting, RAIN achieves $O(\log^{3}(1/\epsilon)/\epsilon^2)$ oracle complexity instead of our $O(\log^2(1/\epsilon)/\epsilon^2)$ rate. RAIN employs a recursive regularization scheme similar to SGD3~\citep{allen2018make}  (inheriting both the $\log^3$ term and also a double-loop structure), while we take a different single-loop, single-oracle call path based on iKM. Our method also accommodates general unknown bias, does not require the distance $D=\operatorname{dist}(x_1,\Fix T)$ as an algorithmic input, and obtains a last-iterate guarantee. 
To our best knowledge, no prior method combines one response per iteration, a single loop, and near-optimal sample complexity for a general nonexpansive map under the stochastic oracle model considered here.

\paragraph{Notation.}
Let $\gH$ be a real Hilbert space with inner product $\ip{\cdot}{\cdot}$,
induced norm $\norm{\cdot}$, and identity operator~$I$. 
The symbols $\E[\cdot]$ and $\E[\,\cdot\mid\gF\,]$ denote expectation and
conditional expectation with respect to a $\sigma$-algebra $\gF$,
respectively; ``a.s.'' means almost surely.  For real-valued functions $f$
and $g$, we write $f=O(g)$ if $|f|\leq Cg$ eventually for some constant
$C>0$, and $f=\Omega(g)$ if $f\geq cg$ eventually for some constant $c>0$.
We write $f=\Theta(g)$ if both $f=O(g)$ and $f=\Omega(g)$, and $f=o(g)$ if
$|f|/g\to0$.  Finally, $f=\widetilde O(g)$ means that $f=O(g)$ up to multiplicative factors that are polynomial in the logarithms of the relevant asymptotic variables.

\input{stochastic_algorithm}
\input{schedule_and_rates}

\section{Conclusion}

Our proposed iKM schedule achieves accelerated last-iterate convergence
for iKM under nonexpansiveness and standard stochastic-oracle assumptions. It reaches both 
near-optimal stochastic sample complexity and the optimal deterministic last-iterate rate, using a simple single-loop update
without variance reduction or recursive regularization.

These guarantees suggest two complementary directions for future work.  First,
it would be valuable to develop and evaluate practical implementations of stochastic iKM in large-scale stochastic first-order and operator-splitting methods. In particular, 
deep neural network training provide an important testbed of how far the benefits extend
beyond the present theory. Second, the fixed-point viewpoint makes it possible to apply the stochastic iKM method for model-free reinforcement learning, where adapting the
inertial mechanism developed here may lead to simple, single-loop model-free
algorithms that are more sample-efficient.

\section*{Acknowledgement}
The work of Y. Chi is supported in part by NSF under DMS-2601955, ECCS-2537189, ECCS-2537078, and AFOSR under FA9550-26-1-B181.

\bibliographystyle{abbrvnat}
\bibliography{references}

\appendix
\input{related_work_appendix}
\input{proofs}

\end{document}

%% file: math_commands.tex
\usepackage{amsmath,amsfonts,bm}
\usepackage{algorithm,algorithmic}

\def\1{\bm{1}}
\DeclareMathAlphabet{\mathsfit}{\encodingdefault}{\sfdefault}{m}{sl}
\SetMathAlphabet{\mathsfit}{bold}{\encodingdefault}{\sfdefault}{bx}{n}
\def\gF{{\mathcal{F}}}

\def\gH{{\mathcal{H}}}

\newcommand{\E}{\mathbb{E}}

\usepackage{amssymb}

%% file: defs.tex
\DeclareMathOperator*{\Fix}{{Fix}}

\usepackage{xcolor}
\definecolor{mydarkblue}{rgb}{0,0.08,0.45}
\definecolor{mygreen}{rgb}{0.032, 0.6392, 0.2039}
\definecolor{mypurple}{HTML}{B266FF}

%% file: stochastic_algorithm.tex
\section{Stochastic iKM under Nonexpansiveness}
\label{sec:stochastic_ikm}

\subsection{The stochastic iKM algorithm}
\label{sec:stochastic_ikm_alg}

Let \(T:\gH\to\gH\) be \emph{nonexpansive}, namely
\begin{equation}\label{eq:stochastic-nonexpansiveness}
 \norm{Tx-Ty}\leq\norm{x-y},
 \qquad \forall x,y\in\gH.
\end{equation}
We also assume its set of fixed points is non-empty, i.e. 
$$\Fix T\coloneqq\{x\in\gH:Tx=x\}\neq\varnothing.$$ 
Fix a horizon \(K\geq2\).
Let the initial points
\(x_0,x_1\in\gH\) be arbitrary and deterministic. The stochastic iKM scheme forms two inertial
extrapolations and performs the oracle-based updates
\begin{equation}\label{eq:sikm}
\begin{aligned}
 y_k&=x_k+\alpha_k(x_k-x_{k-1}),\\
 z_k&=x_k+\beta_k(x_k-x_{k-1}),\\
 x_{k+1}&=(1-\lambda_k)y_k+\lambda_k\widehat T_k(z_k),
 \qquad k=1,\ldots,K.
\end{aligned}
\end{equation}
Here, \(\alpha_k \geq 0\) and \(\beta_k\geq 0\) are the two inertial
extrapolation coefficients associated with \(y_k\) and \(z_k\), respectively,
while \(\lambda_k \in (0,1] \) is the relaxation parameter. Moreover, $\widehat{T}_k$ is a noisy realization of $\widehat{T}_k$, evaluated by calling some oracle model. 

\paragraph{Special case: deterministic iKM.}
When the oracle is exact, so that \(b_k=\xi_k=0\) almost surely and
\(\widehat T_k(z_k)=Tz_k\) for every \(1\leq k\leq K\),
\eqref{eq:sikm} reduces to the deterministic finite-horizon iKM scheme
\begin{equation}\label{eq:deterministic-ikm-update}
\begin{aligned}
 y_k&=x_k+\alpha_k(x_k-x_{k-1}),\\
 z_k&=x_k+\beta_k(x_k-x_{k-1}),\\
 x_{k+1}&=(1-\lambda_k)y_k+\lambda_kTz_k,
 \qquad k=1,\ldots,K.
\end{aligned}
\end{equation}
The two-extrapolation form is the general
inertial Mann iteration of \citet{dong2018general}.  Setting
\(\alpha_k=\beta_k\) makes \(y_k=z_k\) and recovers the common-extrapolation
iKM scheme of \citet{mainge2008convergence}.

%% file: schedule_and_rates.tex
\subsection{Stochastic oracle model and schedule}
\label{sec:one-state-optimistic-schedule}

\paragraph{Stochastic oracle model.} We model the stochastic oracle response in
\eqref{eq:sikm} as for every $k$:
\begin{equation}\label{eq:biased-T-oracle}
 \widehat T_k(z_k)=Tz_k+b_k+\xi_k,
\end{equation}
where \(b_k\) is a predictable, possibly history-dependent bias and \(\xi_k\)
is conditionally centered stochastic noise. 

Consider a probability space equipped with a filtration
\((\gF_k)_{k=1}^K\), where \(\gF_k\subseteq\gF_{k+1}\). For
\(1\leq k\leq K-1\), \(\gF_k\) contains all information available immediately
before the oracle call for update \(k\). All \(\gH\)-valued random variables
are understood to be strongly measurable. For each
\(1\leq k\leq K-1\), the algorithmic quantities
\(x_{k-1},x_k,y_k,z_k\) and the bias \(b_k\) are \(\gF_k\)-measurable, while
\(\xi_k\) (and hence \(\widehat T_k(z_k)\)) is
\(\gF_{k+1}\)-measurable. We make the following assumption regarding the bias and noise variance. 

\begin{itemize}
    \item \textbf{Bias.} The accumulated root-mean-square bias is defined as
    %todo{bounded as $O(\sqrt{K})$?}
\begin{equation}\label{eq:biased-BK}
 B_K\coloneqq\sum_{k=1}^{K-1}
 \left(\E\norm{b_k}^2\right)^{1/2}
\end{equation}
and bounded by $O(\sqrt{K})$.
\item \textbf{Variance.} The noise $\xi_k$ satisfies for some finite constant \(\sigma\geq0\):
\begin{equation}\label{eq:biased-centered-conditions}
 \E[\xi_k\mid\gF_k]=0,
 \qquad
\E[\norm{\xi_k}^2\mid\gF_k]\leq\sigma^2
 \quad\text{a.s.},
 \qquad 1\leq k\leq K.
\end{equation}
\end{itemize}
% Accordingly, whenever the oracle response is integrable, its conditional mean
% is \(Tz_k+b_k\).  No independence across iterations is assumed: the centered
% noises may be adaptively dependent, and only the martingale-difference and
% conditional second-moment conditions in
% \eqref{eq:biased-centered-conditions} are imposed.

\paragraph{Softplus-parameterized iKM schedule.}
For \(K\geq2\), we propose an iKM parameter schedule with one tuning
parameter \(a\) satisfying
\begin{equation}
 0<a\leq \frac{K\log 2}{\log K}-1.
 \label{eq:exact-smooth-a-range}
\end{equation}
Define the auxiliary softplus sequence
\((\omega_k)_{k=0}^K\) by
\begin{equation}
 \omega_k=\omega_k(a)
 \coloneqq \log\!\left(1+K^{a-(a+1)k/K}\right),
 \qquad 0\leq k\leq K,
 \label{eq:exact-smooth-omega}
\end{equation}
For \(1\leq k\leq K\), define %the derived ratio \(r_k\) and transport weight \(u_k\) by
\begin{equation}
 r_k=r_k(a)\coloneqq 2\left(\frac{\omega_{k-1}}{\omega_k}-1\right),
 \qquad
 u_k=u_k(a)\coloneqq \frac{3r_k}{2(1+r_k)(2+r_k)}.
 \label{eq:exact-smooth-r}
\end{equation}
Set the iKM parameters as follows:
\begin{equation}
 \begin{aligned}
 \beta_1&=0,\qquad
 \beta_k=\frac4{3r_{k-1}},\qquad2\leq k\leq K,\\[1mm]
 \lambda_k&=\frac{Kr_k\omega_{k-1}}
 {4(1+1/a)\omega_0}\,u_k,\qquad
 \alpha_k=\frac{u_k-\lambda_k}{1-\lambda_k}\,\beta_k,
 \qquad1\leq k\leq K.
 \end{aligned}
 \label{eq:exact-smooth-ikm-parameters}
\end{equation}

While the schedule may seem abstract at a first glance, the following lemma characterizes the properties of the proposed iKM schedule, whose proof is deferred in Appendix~\ref{app:proof-exact-smooth-schedule}.
\begin{lemma}[Properties of the proposed iKM schedule]
\label{lem:exact-smooth-schedule-properties}
For \(K\geq2\) and \(a\) satisfying
\eqref{eq:exact-smooth-a-range}, the parameters in
\eqref{eq:exact-smooth-ikm-parameters} satisfy
\begin{equation}
 \beta_k>0\quad(2\leq k\leq K),
 \qquad
 0<\lambda_k<1,
 \qquad
 0\leq\alpha_k<1
 \quad(1\leq k\leq K).
 \label{eq:exact-smooth-parameter-ranges}
\end{equation}
In addition,
\begin{enumerate}[label=\textup{(\alph*)}]
 \item \(\alpha_1=0\) and \(\alpha_k=\Theta(1)\) for
 \(2\leq k\leq K\).

 \item \(\beta_2>\cdots>\beta_K\), with
 \begin{equation}
  \begin{aligned}
   \beta_2
   =\Theta\!\left(\frac{K}{1+a}
   \left(a+\frac1{\log K}\right)\right),
   \qquad
   \beta_K=\Theta\!\left(\frac{K}{(1+a)\log K}\right).
  \end{aligned}
  \label{eq:exact-smooth-beta-endpoint-orders}
 \end{equation}

 \item \((\lambda_k)_{k=1}^K\) is unimodal.  Its
 maximizing set consists of one index or two adjacent indices; the sequence
 is strictly increasing before this set and strictly decreasing after it.
 If \(k_\lambda\) is any maximizing index, then
 \begin{equation}
  k_\lambda
  =\max\!\left\{1,
  \frac{aK}{a+1}
  -\frac{\log(2\rho_\lambda)}{a+1}\frac{K}{\log K}\right\}+O(1).
 \label{eq:exact-smooth-lambda-peak-location}
 \end{equation}
 Here \(\rho_\lambda=1.2564312\ldots\) is the unique positive solution of
 \(e^\rho=1+2\rho\), and
 \begin{equation}
  \begin{aligned}
   \lambda_1
   &=\Theta\!\left(
   \frac{a(1+a)\log^2 K}
   {K(1+a\log K)^2}
   \right), \quad
   \max_{1\leq k\leq K}\lambda_k
   &=\Theta(1+a\log K)\,\lambda_1,
   \quad
\lambda_K=\Theta\!\left(\frac{1+a\log K}{K}\right)\lambda_1.
  \end{aligned}
  \label{eq:exact-smooth-lambda-endpoint-orders}
 \end{equation}
\end{enumerate}
\end{lemma}

\begin{figure}[ht]
 \centering
 \begin{tabular}{@{}c@{}c@{}}
\includegraphics[width=0.49\linewidth]{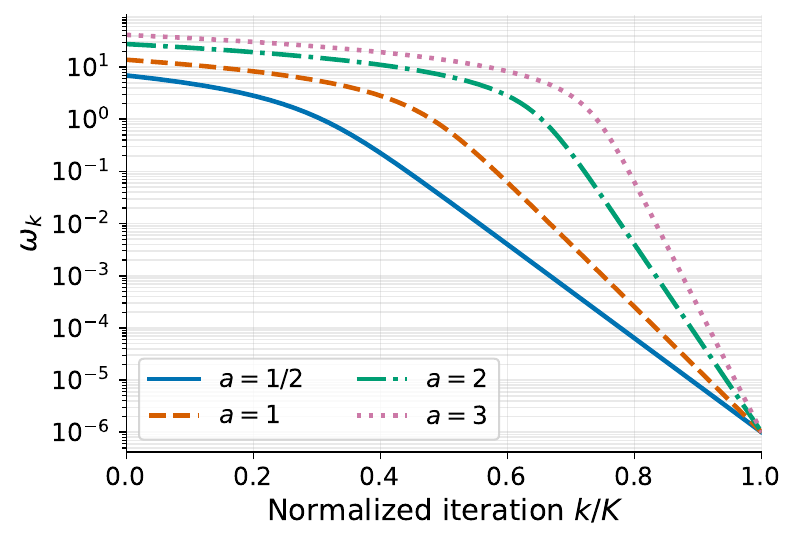}
&
\includegraphics[width=0.49\linewidth]{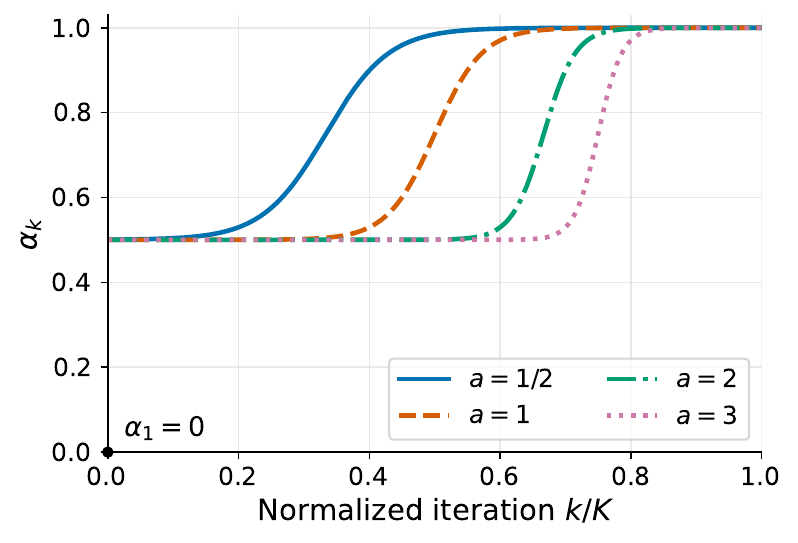} \\
   {\(\omega_k\)} & {\(\alpha_k\)} \\[1mm]
\includegraphics[width=0.49\linewidth]{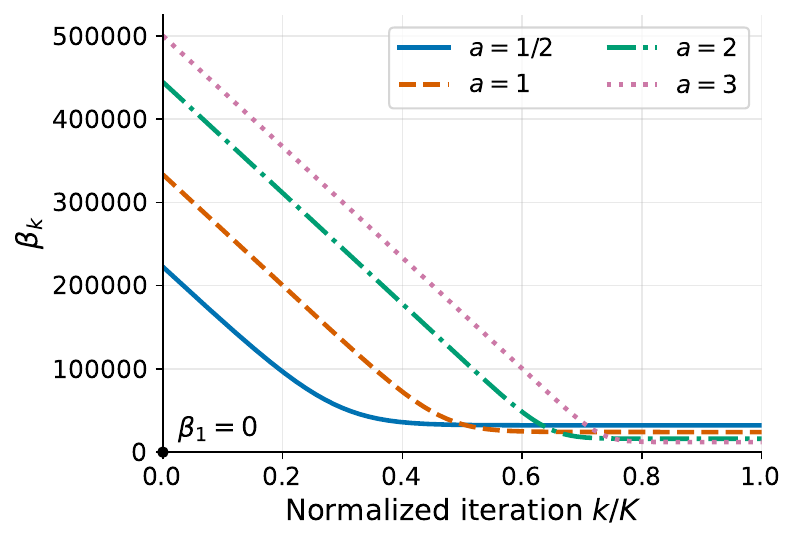}
&
\includegraphics[width=0.49\linewidth]{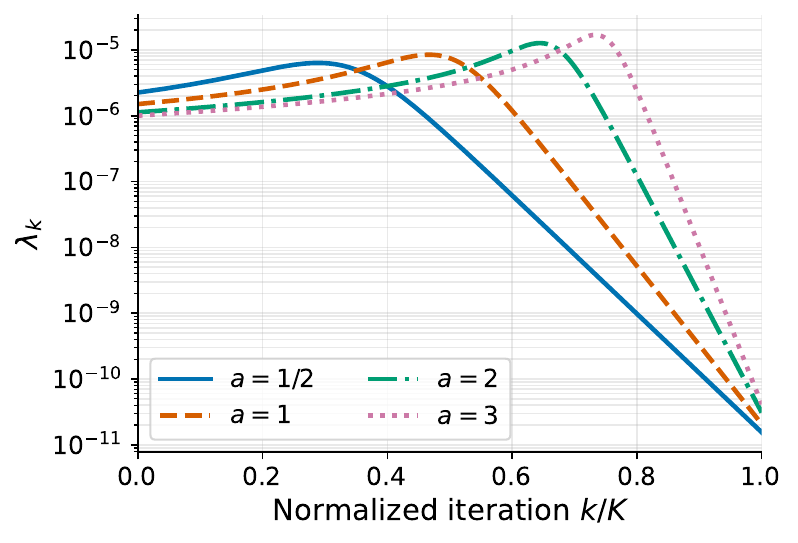} \\
   {\(\beta_k\)} & {\(\lambda_k\)}
\end{tabular}
 \caption{The auxiliary softplus sequence from \eqref{eq:exact-smooth-omega} and
 iKM parameters from \eqref{eq:exact-smooth-ikm-parameters} for \(K=10^6\) and
 \(a\in\{1/2,1,2,3\}\).  The horizontal axis is the normalized iteration
 \(k/K\).  The isolated values
 \(\alpha_1=\beta_1=0\) are marked separately.  The \(\omega_k\) and
 \(\lambda_k\) panels use logarithmic vertical scales.}
 \label{fig:exact-smooth-ikm-parameters}
\end{figure}

Figure~\ref{fig:exact-smooth-ikm-parameters} illustrates the auxiliary
softplus sequence and the three hyperparameter sequences.  Increasing \(a\) delays the transition of \(\alpha_k\) towards $1$, and shifts the peak of \(\lambda_k\) toward the end of the horizon, while 
\(\beta_k\) decreases for \(k\geq2\) after the large initial
value \(\beta_1=0\).

\paragraph{Convergence analysis.}
 We are now ready to present the convergence rate of stochastic iKM.

\begin{theorem}[Convergence rate for stochastic iKM]
\label{thm:onestate-terminal-rate}
Under the filtration and measurability setting of
Subsection~\ref{sec:stochastic_ikm_alg}, suppose that \(\Fix T\ne\varnothing\),
\(T\) satisfies nonexpansiveness \eqref{eq:stochastic-nonexpansiveness}, and
\eqref{eq:biased-T-oracle}, \eqref{eq:biased-centered-conditions} hold.
Fix \(K\geq2\), keep the initial points \(x_0,x_1\) arbitrary and deterministic,
choose \(a\) satisfying \eqref{eq:exact-smooth-a-range}, and generate
\(x_K,y_K,z_K\) by \eqref{eq:sikm} using the schedule
\eqref{eq:exact-smooth-ikm-parameters}.  Then there is an absolute constant
\(C>0\) such that
\begin{align}
 &\max_{v_K\in\{x_K,y_K,z_K\}}
 \left(\E\norm{v_K-Tv_K}^2\right)^{1/2}\notag\\
 &\leq C\left\{
 \frac{1+1/a}{K}
 \inf_{p^\star\in\Fix T}\norm{x_1-p^\star}
 +(1+a)\left(
   \frac{\sigma\log K}{\sqrt K}
   +\frac{B_K\log K}{K}
  \right)\right\}.
 \label{eq:onestate-terminal-rate}
\end{align}
\end{theorem}
The proof is in Appendix~\ref{app:proof_stochastic_rate}. For every fixed \(a=\Theta(1)\) satisfying \eqref{eq:exact-smooth-a-range}, the first term in
\eqref{eq:onestate-terminal-rate} characterizes the deterministic contribution, while the centered-noise
contribution is encapsulated in the second term. The dependence on the tuning parameter \(a\) displays a mild
problem-dependent trade-off: increasing \(a\) improves the
deterministic factor \(1+1/a\) but worsens the noise and bias factor \(1+a\),
whereas decreasing \(a\) has the opposite effect. The convergence rate simultaneously achieves the optimal rates for both deterministic and stochastic settings under our proposed schedule.

\paragraph{Optimal sample complexity of stochastic iKM.}
Suppose \(B_K\leq B\sqrt K\) for some constant \(B\geq0\), and fix a constant
\(a=\Theta(1)\) satisfying \eqref{eq:exact-smooth-a-range}. 
Since each iteration uses one oracle call,
Theorem~\ref{thm:onestate-terminal-rate} implies that, for small enough \(\epsilon>0\), it suffices to take
\begin{equation}
 \begin{aligned}
 K=O\!\Bigg(\frac{1}{\epsilon}
    \inf_{p^\star\in\Fix T}\norm{x_1-p^\star}+\frac{(\sigma+B)^2}{\epsilon^2}
    \log^2\!\left(\frac{\sigma+B}{\epsilon}\right)
 \Bigg)
 \end{aligned}
 \label{eq:onestate-sample-complexity}
\end{equation}
to guarantee
\begin{equation}
 \max_{v_K\in\{x_K,y_K,z_K\}}
 \left(\E\norm{v_K-Tv_K}^2\right)^{1/2}\leq\epsilon.
\end{equation}

Notably, in the unbiased case \(b_k\equiv0\) (so \(B=0\)), with
\(\sigma>0\), \eqref{eq:onestate-sample-complexity} matches the stochastic
lower bound inherited from smooth convex optimization
\citep[Theorem~2]{foster2019complexity} up to one logarithmic factor, while
improving the dependence on \(\epsilon\) relative to the
\(O(\epsilon^{-4})\) random-iterate bound known for stochastic KM under a
related additive martingale-difference oracle model
\citep[Proposition~5.1 and Corollary~5.4]{bravo2024stochastic}.

\paragraph{Optimal convergence rate of deterministic iKM under nonexpansiveness.} In the deterministic setting where $\sigma=0$ and there's no bias, Theorem~\ref{thm:onestate-terminal-rate} implies that the deterministic iKM under nonexpansiveness achieves $O(1/K)$ last-iterate rate, which is
\emph{worst-case optimal}~\citep[Theorem~4.6]{park2022exact}.  
% Indeed, \citet[Theorem~4.6]{park2022exact} proved that,
% for any deterministic
% fixed-point method and any horizon $k$, there exists a nonexpansive operator
% $T$ with a fixed point $p^\star$ such that the generated iterate $u_k$
% satisfies
% \[
%         \norm{u_k-Tu_k}
%         \geq
%         \frac{2}{k+1}\norm{u_0-p^\star}.
% \]
Moreover, this rate strictly improves the slower $O(1/\sqrt{K})$ rate of the no-inertia KM iteration
\eqref{eq:km_iteration}~\citep{davis2016convergence}.

%% file: related_work_appendix.tex
% \section{More Discussion on Related Works}
% \label{app:more-related-work}

\section{Relation to Fast KM}
\label{app:relation-fast-km}

Fast KM was introduced by \citet{bot2023fast} as an accelerated method for finding
fixed points of nonexpansive operators.  A recent generalization provides an
explicit nonasymptotic $O(K^{-1})$ fixed-point-residual bound
\citep[Remark~5]{bot2026overrelaxation}, matching the optimal worst-case order
for general nonexpansive fixed-point problems
\citep[Theorem~4.6]{park2022exact}.

We first show that Fast KM can be viewed as iKM with specific parameters. Following the notation of Fast KM, let $\alpha>2$ and $0<s\leq1$.  Starting
from arbitrary $z_0,z_1\in\gH$, Fast KM performs, for every $k\geq1$, the
iteration
\begin{equation}
 \begin{aligned}
 z_{k+1}
 &=\left(1-\frac{s\alpha}{2(k+\alpha)}\right)z_k
   +\frac{(1-s)k}{k+\alpha}(z_k-z_{k-1})
   +\frac{s\alpha}{2(k+\alpha)}Tz_k
   +\frac{sk}{k+\alpha}(Tz_k-Tz_{k-1}).
 \end{aligned}
 \label{eq:query-fast-km-recurrence}
\end{equation}
See \citet[Algorithm~2.1]{bot2023fast} and
\citet[Section~3.3.1, Eq.~(3.21)]{bot2026overrelaxation}.  To express
\eqref{eq:query-fast-km-recurrence} as iKM, choose iKM parameters
\begin{equation}
 \alpha_k=\frac{k(1-s)}{k+\alpha-\frac{s}{2}(\alpha-2)},
 \qquad
 \beta_k=\frac{2k}{\alpha-2},
 \qquad
 \lambda_k=\frac{s(\alpha-2)}{2(k+\alpha)}.
 \label{eq:ikm-fast-km-parameters}
\end{equation}

\begin{proposition}[Fast KM as deterministic iKM]
\label{prop:ikm-query-is-fast-km}
Let $z_0,z_1\in\gH$ be arbitrary Fast-KM initial points.  Run deterministic
iKM \eqref{eq:deterministic-ikm-update} with the parameters in
\eqref{eq:ikm-fast-km-parameters} and the base points
\begin{equation}
 \begin{aligned}
 x_0&=2z_1-(1-s)z_0-sTz_0,\\
 x_1&=\left(1+\frac{2}{\alpha}\right)z_1
      -\frac{2(1-s)}{\alpha}z_0-\frac{2s}{\alpha}Tz_0.
 \end{aligned}
 \label{eq:app-fast-ikm-initialization}
\end{equation}
Then its query sequence satisfies \eqref{eq:query-fast-km-recurrence} for
every $k\geq1$. 
\end{proposition}
The proof of Proposition \ref{prop:ikm-query-is-fast-km} is given in Appendix~\ref{app:proof-fast-km-ikm}. It is worth highlighting that the parameters in
\eqref{eq:ikm-fast-km-parameters} differ from those in our main schedule~\eqref{eq:exact-smooth-ikm-parameters}. While 
Fast KM achieve the optimal rate under exact deterministic evaluations, our schedule \eqref{eq:exact-smooth-ikm-parameters} is optimal for both deterministic and
stochastic regimes.  

\subsection{Proof of Proposition \ref{prop:ikm-query-is-fast-km}}\label{app:proof-fast-km-ikm}

Let
\begin{align}\label{eq:app-fast-d-definition}
 d_k\coloneqq x_k-x_{k-1},
 \qquad
 G\coloneqq I-T.
\end{align}
The initialization \eqref{eq:app-fast-ikm-initialization} gives
\begin{equation}
 d_1=-\frac{\alpha-2}{\alpha}
 \bigl(z_1-(1-s)z_0-sTz_0\bigr).
 \label{eq:app-fast-initial-difference}
\end{equation}
Since $\beta_1=2/(\alpha-2)$ by \eqref{eq:ikm-fast-km-parameters}, from the deterministic iKM scheme \eqref{eq:deterministic-ikm-update} and \eqref{eq:app-fast-ikm-initialization} we know that the first iKM query is indeed
\begin{align*}
 x_1+\beta_1d_1
 &=z_1+\frac{2}{\alpha}
   \bigl(z_1-(1-s)z_0-sTz_0\bigr)
   -\frac{2}{\alpha}
   \bigl(z_1-(1-s)z_0-sTz_0\bigr)
 =z_1.
\end{align*}

From
\eqref{eq:deterministic-ikm-update} and the identity
$z_k=x_k+\beta_kd_k$, we have
\begin{align}
 x_{k+1}
 &=(1-\lambda_k)(x_k+\alpha_kd_k)+\lambda_kTz_k\notag\\
 &\overset{\eqref{eq:app-fast-d-definition}}= (1-\lambda_k)(x_k+\alpha_kd_k)
   +\lambda_k(z_k-Gz_k)\notag\\
 &=x_k+\bigl((1-\lambda_k)\alpha_k+\lambda_k\beta_k\bigr)d_k
   -\lambda_kGz_k.
 \label{eq:app-fast-ikm-expansion}
\end{align}

The parameter choice \eqref{eq:ikm-fast-km-parameters} yields
\[
 (1-\lambda_k)\alpha_k
 =\frac{k+\alpha-\frac{s}{2}(\alpha-2)}{k+\alpha}
  \frac{k(1-s)}{k+\alpha-\frac{s}{2}(\alpha-2)}
 =\frac{k(1-s)}{k+\alpha},
\]
and
\[
 \lambda_k\beta_k
 =\frac{s(\alpha-2)}{2(k+\alpha)}\frac{2k}{\alpha-2}
 =\frac{sk}{k+\alpha}.
\]
Consequently,
\begin{equation}
 (1-\lambda_k)\alpha_k+\lambda_k\beta_k
 =\frac{k}{k+\alpha}.
 \label{eq:app-fast-inertial-coefficient}
\end{equation}
Using \eqref{eq:app-fast-inertial-coefficient} in
\eqref{eq:app-fast-ikm-expansion} gives
\begin{equation}
 d_{k+1}=\frac{k}{k+\alpha}d_k
 -\frac{s(\alpha-2)}{2(k+\alpha)}Gz_k.
 \label{eq:app-fast-d-recurrence}
\end{equation}

The identity $z_j=x_j+\beta_jd_j$ further gives
\begin{align}
 z_{k+1}-z_k
 &=(1+\beta_{k+1})d_{k+1}-\beta_kd_k\notag\\
 &\overset{\eqref{eq:app-fast-d-recurrence}}=\left((1+\beta_{k+1})\frac{k}{k+\alpha}-\beta_k\right)d_k
   -(1+\beta_{k+1})\frac{s(\alpha-2)}{2(k+\alpha)}Gz_k.
 \label{eq:app-fast-z-increment-preliminary}
\end{align}
By \eqref{eq:ikm-fast-km-parameters},
\[
 (1+\beta_{k+1})\frac{k}{k+\alpha}-\beta_k
 =\frac{2k+\alpha}{\alpha-2}\frac{k}{k+\alpha}
  -\frac{2k}{\alpha-2}
 =-\frac{\alpha k}{(\alpha-2)(k+\alpha)},
\]
and
\[
 (1+\beta_{k+1})\frac{s(\alpha-2)}{2(k+\alpha)}
 =\frac{2k+\alpha}{\alpha-2}
  \frac{s(\alpha-2)}{2(k+\alpha)}
 =\frac{s(2k+\alpha)}{2(k+\alpha)}.
\]
Substituting these identities into
\eqref{eq:app-fast-z-increment-preliminary} yields
\begin{equation}
 z_{k+1}-z_k
 =-\frac{\alpha k}{(\alpha-2)(k+\alpha)}d_k
  -\frac{s(2k+\alpha)}{2(k+\alpha)}Gz_k.
 \label{eq:app-fast-z-increment}
\end{equation}

At $k=1$, \eqref{eq:app-fast-initial-difference} and
\eqref{eq:app-fast-z-increment} give
\begin{align*}
 z_2-z_1
 &=\frac{z_1-(1-s)z_0-sTz_0}{1+\alpha}
   -\frac{s(2+\alpha)}{2(1+\alpha)}(z_1-Tz_1).
\end{align*}
Rearranging this identity gives
\begin{align*}
 z_2
 &=\left(1-\frac{s\alpha}{2(1+\alpha)}\right)z_1
   +\frac{1-s}{1+\alpha}(z_1-z_0)+\frac{s\alpha}{2(1+\alpha)}Tz_1
   +\frac{s}{1+\alpha}(Tz_1-Tz_0),
\end{align*}
which is \eqref{eq:query-fast-km-recurrence} at $k=1$.

It remains to verify \eqref{eq:query-fast-km-recurrence} for $k\geq2$.
\eqref{eq:app-fast-z-increment} and
\eqref{eq:app-fast-d-recurrence} at index $k-1$ give
\begin{align*}
 z_k-z_{k-1}
 &=-\frac{\alpha(k-1)}{(\alpha-2)(k-1+\alpha)}d_{k-1}
   -\frac{s(2k-2+\alpha)}{2(k-1+\alpha)}Gz_{k-1},\\
 d_k
 &=\frac{k-1}{k-1+\alpha}d_{k-1}
   -\frac{s(\alpha-2)}{2(k-1+\alpha)}Gz_{k-1}.
\end{align*}
Eliminating $d_{k-1}$ yields
\begin{equation}
 d_k=-\frac{\alpha-2}{\alpha}(z_k-z_{k-1})
     -\frac{s(\alpha-2)}{\alpha}Gz_{k-1},
 \qquad k\geq2.
 \label{eq:app-fast-d-elimination}
\end{equation}
Substituting \eqref{eq:app-fast-d-elimination} into
\eqref{eq:app-fast-z-increment}, we obtain
\begin{align}
 z_{k+1}-z_k
 &=\frac{k}{k+\alpha}(z_k-z_{k-1})
   +\frac{sk}{k+\alpha}Gz_{k-1}
   -\frac{s(2k+\alpha)}{2(k+\alpha)}Gz_k\notag\\
 &=\frac{k}{k+\alpha}(z_k-z_{k-1})
   -\frac{s\alpha}{2(k+\alpha)}Gz_k
   -\frac{sk}{k+\alpha}(Gz_k-Gz_{k-1}).
 \label{eq:app-fast-query-G-form}
\end{align}
Finally, expanding $G=I-T$ in \eqref{eq:app-fast-query-G-form} gives
\begin{align*}
 z_{k+1}
 &=\left(1-\frac{s\alpha}{2(k+\alpha)}\right)z_k
   +\frac{(1-s)k}{k+\alpha}(z_k-z_{k-1})
   +\frac{s\alpha}{2(k+\alpha)}Tz_k
   +\frac{sk}{k+\alpha}(Tz_k-Tz_{k-1}),
\end{align*}
which is \eqref{eq:query-fast-km-recurrence} for every $k\geq2$.  Together
with the $k=1$ identity above, this proves the claimed representation.

%% file: proofs.tex
\section{Proofs for stochastic iKM}
\label{app:one-state-optimistic-schedule}

\subsection{Auxiliary Lemmas}\label{app:auxiliary-lemmas}

Define the residual operator
\begin{equation}
 G\coloneqq I-T.
 \label{eq:app-residual-operator}
\end{equation}
At this point we need the following
consequences of nonexpansiveness, whose proof is deferred to Appendix~\ref{app:proof-residual-operator}.
\begin{lemma}[Residual operator of a nonexpansive map]
\label{lem:app-residual-operator}
Suppose that \(T:\gH\to\gH\) is nonexpansive.  Then \(G\) is
\(1/2\)-cocoercive, hence monotone, and \(2\)-Lipschitz:
\begin{equation}
 \forall x,y\in\gH,\quad\frac12\norm{Gx-Gy}^2
 \leq \ip{Gx-Gy}{x-y},
 \qquad
 \norm{Gx-Gy}\leq2\norm{x-y}.
 \label{eq:app-residual-operator-properties}
\end{equation}
Moreover, the zeros of \(G\) are exactly the fixed points of \(T\):
\begin{equation}
 Gp^\star=0\quad\Longleftrightarrow\quad p^\star\in\Fix T.
 \label{eq:app-residual-zero-fixed-point}
\end{equation}
\end{lemma}

We will also use the following regularized-root lemma repeatedly. The proof is in Appendix~\ref{app:proof-existence}.

\begin{lemma}[Existence and firm nonexpansiveness]
\label{lem:existence}
Let \(G:\gH\to\gH\) be monotone and continuous.  For every \(\mu>0\) and
\(v\in\gH\), the equation
\begin{equation}
 Gp+\mu(p-v)=0
\end{equation}
has a unique solution.  If \(p_\mu(v)\) denotes this solution, then the
solution map is firmly nonexpansive: for all \(v,v'\in\gH\),
\begin{equation}
 \norm{p_\mu(v)-p_\mu(v')}^2
 \leq\ip{p_\mu(v)-p_\mu(v')}{v-v'}.
 \label{eq:regularized-root-firm-nonexpansiveness}
\end{equation}
\end{lemma}

\paragraph{Auxiliary parameters.}
Define
\begin{equation}
 \mu_k=
 \frac{4(1+1/a)\omega_0}{K\omega_{k-1}}>0,
 \qquad 1\leq k\leq K+1,
 \label{eq:app-exact-proof-masses}
\end{equation}
where \(\omega_k\) is defined in \eqref{eq:exact-smooth-omega}.
Then by \eqref{eq:app-exact-proof-masses}, \eqref{eq:exact-smooth-r}, and
\eqref{eq:exact-smooth-ikm-parameters} we have
\begin{equation}
  \forall 1\leq k\leq K:\quad \frac{\mu_{k+1}}{\mu_k}
 =\frac{\omega_{k-1}}{\omega_k}
 =1+\frac{r_k}{2},
 \qquad
 \lambda_k=u_k\frac{r_k}{\mu_k}.
 \label{eq:app-exact-schedule-conversion}
\end{equation}
We will use the following basic bounds for these schedule quantities repetitively. The proof is deferred to Appendix~\ref{app:proof-exact-schedule-bounds}.
\begin{lemma}[Auxiliary bounds for the schedule]
\label{lem:app-exact-schedule-bounds}
For \(K\geq2\) and \(a\) satisfying
\eqref{eq:exact-smooth-a-range}, consider the schedule
\eqref{eq:exact-smooth-omega}--\eqref{eq:exact-smooth-ikm-parameters}.
Then, for \(1\leq k\leq K\),
\begin{equation}
 1<\frac{\mu_{k+1}}{\mu_k}
 =\frac{\omega_{k-1}}{\omega_k}
 <K^{(a+1)/K}\leq2.
 \label{eq:app-exact-clock-ratio-bound}
\end{equation}
Moreover, for \(1\leq k\leq K\),
\begin{subequations}
\begin{align}
 0<r_k&<2,
 \label{eq:r_k_in_(0,2)}\\
 r_k&<\frac{2(a+1)K^{(a+1)/K}\log K}{K},
 \label{eq:app-exact-clock-r-uniform}\\
 0<\frac{r_k}{\mu_k}&<\frac56,
 \label{eq:app-exact-clock-step-ratio-uniform-gap}\\
 0<\lambda_k&<u_k<1.
 \label{eq:app-exact-lambda-range}
\end{align}
\end{subequations}
For \(2\leq k\leq K\),
\begin{equation}
 \begin{gathered}
  \beta_k>0,\qquad 0<\frac{\alpha_k}{\beta_k}
  =\frac{u_k-\lambda_k}{1-\lambda_k}
  =u_k\frac{\mu_k-r_k}{\mu_k-u_kr_k}
  <u_k<1.
 \end{gathered}
 \label{eq:app-exact-alpha-ratio}
\end{equation}
\end{lemma}

\subsection{Proof of
Lemma~\ref{lem:exact-smooth-schedule-properties}}
\label{app:proof-exact-smooth-schedule}

We first prove the parameter ranges and part~\textup{(a)}.
Steps 2--4 establish parts~\textup{(b)} and~\textup{(c)}.

\paragraph{Step 1: parameter ranges and the order of \(\alpha_k\).}
Define the continuous interpolation of the auxiliary
sequence \((\omega_k)_{k=0}^K\) from \eqref{eq:exact-smooth-omega} on
\([0,1]\) by
\begin{equation}
 \omega(t)=\log(1+K^{a-(a+1)t}),
 \qquad
 z(t)=K^{a-(a+1)t},
 \qquad 0\leq t\leq1.
 \label{eq:app-exact-continuous-clock}
\end{equation}
Then
\begin{align}\label{eq:omega_discrete_continuous}
\omega(k/K)=\omega_k,\qquad \forall 0\leq k\leq K.
\end{align}
It follows that \(\omega\) is strictly decreasing since
%   Because
% \(z'(t)=-(a+1)(\log K)z(t)\), the relation
% \(\omega(t)=\log(1+z(t))\) in
% \eqref{eq:app-exact-continuous-clock} yields
\begin{equation}
 \omega'(t)=-(a+1)\log K\,\frac{z(t)}{1+z(t)}<0.
 \label{eq:app-exact-clock-first-derivative}
\end{equation}
% Equation \eqref{eq:app-exact-continuous-clock} gives \(\omega(t)>0\), and

Multiplying \eqref{eq:app-exact-alpha-ratio} by \(\beta_k\) yields the
factorization
\begin{equation}
 \alpha_k=(u_k\beta_k)
 \frac{\mu_k-r_k}{\mu_k-u_kr_k},
 \qquad 2\leq k\leq K.
 \label{eq:app-exact-alpha-factorization}
\end{equation}
Thus it remains to bound the two factors on the right-hand side.

For the first factor, the formulas for \(u_k\) and \(\beta_k\) in
\eqref{eq:exact-smooth-r} and
\eqref{eq:exact-smooth-ikm-parameters} give
\begin{equation}
 u_k\beta_k
 =\frac{2r_k}{r_{k-1}(1+r_k)(2+r_k)},
 \qquad 2\leq k\leq K.
 \label{eq:app-exact-alpha-product-identity}
\end{equation}
Its lower bound requires \(r_{k-1}<r_k\).  To prove this strict
monotonicity, define
\begin{equation}
 F(t)=-\log\omega(t),
 \qquad
 \phi(s)=\frac{1-e^{-s}}s,
 \qquad s>0.
 \label{eq:app-exact-clock-phi}
\end{equation}
Since \(e^s>1+s\) for \(s>0\),
\begin{equation}
 \phi'(s)=\frac{(s+1)e^{-s}-1}{s^2}<0.
 \label{eq:phi'}
\end{equation}
Furthermore,
\begin{equation}
 F'(t)=(a+1)\log K\,\phi(\omega(t)),
 \qquad
 F''(t)=(a+1)\log K\,\phi'(\omega(t))\omega'(t)>0,
 \label{eq:app-exact-clock-F-derivative}
\end{equation}
where the first identity follows from
\(z(t)/(1+z(t))=1-e^{-\omega(t)}\), which is a consequence of
\eqref{eq:app-exact-continuous-clock}; the second uses \eqref{eq:phi'} and
\eqref{eq:app-exact-clock-first-derivative}.  For \(1\leq k\leq K\),
\begin{equation}
 \log\frac{\mu_{k+1}}{\mu_k}
 =F(k/K)-F((k-1)/K)
 =\int_{(k-1)/K}^{k/K}F'(t)\,dt,
 \label{eq:app-exact-clock-log-ratio-increment}
\end{equation}
where the first equality uses \eqref{eq:app-exact-schedule-conversion} and
\eqref{eq:omega_discrete_continuous}.  Since \(F'\) is strictly increasing by the second relation in \eqref{eq:app-exact-clock-F-derivative},
the integrals in \eqref{eq:app-exact-clock-log-ratio-increment} increase
strictly with \(k\).  The first identity in
\eqref{eq:app-exact-schedule-conversion} and
\eqref{eq:r_k_in_(0,2)} therefore give
\begin{equation}
 0<r_1<r_2<\cdots<r_K<2.
 \label{eq:app-exact-clock-r-monotonicity}
\end{equation}

For the upper bound in \eqref{eq:app-exact-alpha-product-identity}, we also
need a lower bound on \(r_{k-1}\) in terms of \(r_k\).  A second
differentiation of \eqref{eq:app-exact-clock-first-derivative} gives
\begin{equation}
 \omega''(t)=(a+1)^2(\log K)^2
 \frac{z(t)}{(1+z(t))^2}>0.
 \label{eq:app-exact-clock-second-derivative}
\end{equation}
Thus \(\omega'\) is increasing.  For \(2\leq k\leq K\),
\begin{equation}
 \omega_{k-2}-\omega_{k-1}
 =-\int_{(k-2)/K}^{(k-1)/K}\omega'(t)\,dt
 \geq-\int_{(k-1)/K}^{k/K}\omega'(t)\,dt
 =\omega_{k-1}-\omega_k.
 \label{eq:app-exact-clock-delta-monotonicity}
\end{equation}
Consequently, \eqref{eq:app-exact-clock-delta-monotonicity},
\eqref{eq:exact-smooth-r}, and
\eqref{eq:app-exact-schedule-conversion} give
\begin{equation}
 r_{k-1}
 =\frac{2(\omega_{k-2}-\omega_{k-1})}{\omega_{k-1}}
 \geq\frac{2(\omega_{k-1}-\omega_k)}{\omega_{k-1}}
 =\frac{2r_k}{r_k+2},
 \qquad 2\leq k\leq K.
 \label{eq:app-exact-clock-consecutive-r}
\end{equation}
Therefore, for \(2\leq k\leq K\),
\begin{equation}
 \frac16
 <\frac{2}{(1+r_k)(2+r_k)}
 <u_k\beta_k
 \leq\frac1{1+r_k}<1,
 \label{eq:app-exact-clock-alpha-product-order}
\end{equation}
where the first inequality follows from \(r_k<2\) in
\eqref{eq:r_k_in_(0,2)}, the second inequality follows from
\eqref{eq:app-exact-clock-r-monotonicity} and
\eqref{eq:app-exact-alpha-product-identity},
the third inequality follows from \eqref{eq:app-exact-clock-consecutive-r} and
\eqref{eq:app-exact-alpha-product-identity}, and
the final inequality follows from \(r_k>0\) in
\eqref{eq:r_k_in_(0,2)}.

For the second factor in \eqref{eq:app-exact-alpha-factorization},
\eqref{eq:app-exact-clock-step-ratio-uniform-gap} and
\eqref{eq:app-exact-lambda-range} give
\begin{align}\label{eq:factor2_lb}
 \frac{\mu_k-r_k}{\mu_k-u_kr_k}
 =\frac{1-r_k/\mu_k}{1-u_kr_k/\mu_k}>1-\frac{r_k}{\mu_k}.
\end{align}
Thus
\begin{equation}
 \frac16
 \overset{\eqref{eq:app-exact-clock-step-ratio-uniform-gap}}<1-\frac{r_k}{\mu_k}
 \overset{\eqref{eq:factor2_lb}}<\frac{\mu_k-r_k}{\mu_k-u_kr_k}<1,
 \qquad 2\leq k\leq K,
 \label{eq:app-exact-clock-alpha-ratio-order}
\end{equation}
where the final inequality follows because
\eqref{eq:r_k_in_(0,2)} and \eqref{eq:app-exact-lambda-range} give
\[
 (\mu_k-u_kr_k)-(\mu_k-r_k)=r_k(1-u_k)>0.
\]
Combining \eqref{eq:app-exact-alpha-factorization},
\eqref{eq:app-exact-clock-alpha-product-order}, and
\eqref{eq:app-exact-clock-alpha-ratio-order} gives
\begin{equation}
 \frac1{36}<\alpha_k<1,
 \qquad 2\leq k\leq K.
 \label{eq:app-exact-alpha-uniform-order}
\end{equation}
Also, \eqref{eq:exact-smooth-ikm-parameters} gives \(\alpha_1=0\).
This proves \eqref{eq:exact-smooth-parameter-ranges} and part~\textup{(a)}.

\paragraph{Step 2: monotonicity and endpoint orders of \(\beta_k\).}
Since \(\beta_k=4/(3r_{k-1})\) by
\eqref{eq:exact-smooth-ikm-parameters} for all $2\leq k\leq K$,
\eqref{eq:app-exact-clock-r-monotonicity} immediately gives
\begin{equation}
 \beta_2>\beta_3>\cdots>\beta_K>0.
 \label{eq:app-exact-beta-monotonicity}
\end{equation}
It remains to prove the two endpoint orders \eqref{eq:exact-smooth-beta-endpoint-orders}.  We first write each successive
logarithmic ratio in a form that can be estimated locally.
\eqref{eq:app-exact-clock-F-derivative} and
\eqref{eq:app-exact-clock-log-ratio-increment} give
\begin{equation}
 \log\frac{\mu_{k+1}}{\mu_k}
 =\int_{(k-1)/K}^{k/K}(a+1)\log K\,\phi(\omega(t))\,dt.
 \label{eq:app-exact-clock-log-increment-integral}
\end{equation}
To bound the integrand, observe that by \eqref{eq:app-exact-clock-phi}, for every \(s>0\),
\begin{equation}
 \frac1{1+s}\leq\phi(s)\leq\frac2{1+s},
 \label{eq:app-exact-clock-phi-comparison}
\end{equation}
where the lower bound follows from \(e^s\geq1+s\);  for the upper bound,
use \(\phi(s)\leq1\leq2/(1+s)\) when \(s\leq1\), and
\(\phi(s)\leq1/s\leq2/(1+s)\) when \(s\geq1\).
Both \(\omega\) and \(\phi\) are decreasing by
\eqref{eq:app-exact-clock-first-derivative} and \eqref{eq:phi'}, so the integrand in
\eqref{eq:app-exact-clock-log-increment-integral} is increasing and therefore
\[
 \frac{(a+1)\log K}{K}\,\phi(\omega_{k-1})
 \leq\log\frac{\mu_{k+1}}{\mu_k}
 \leq\frac{(a+1)\log K}{K}\,\phi(\omega_k).
\]
Applying \eqref{eq:app-exact-clock-phi-comparison} to these two endpoint
values, and using
\(1+\omega_{k-1}<2(1+\omega_k)\) from
\eqref{eq:app-exact-clock-ratio-bound}, yields
\begin{equation}
 \frac{(a+1)\log K}{2K(1+\omega_k)}
 <\frac{(a+1)\log K}{K(1+\omega_{k-1})}
 \leq\log\frac{\mu_{k+1}}{\mu_k}
 \leq\frac{2(a+1)\log K}{K(1+\omega_k)}.
 \label{eq:app-exact-clock-log-increment-two-sided}
\end{equation}
Apply the elementary inequalities
\begin{align}\label{eq:elementary_log_ineq}
\log q<q-1\leq q\log q \quad \text{for}\quad 1<q<2,
\end{align}
with $q=\frac{\mu_{k+1}}{\mu_k}\in(1,2)$,
where the inclusion follows from
\eqref{eq:app-exact-clock-ratio-bound}.  Using
\(r_k=2(q-1)\) from \eqref{eq:app-exact-schedule-conversion} then gives
\begin{equation}
 2\log\frac{\mu_{k+1}}{\mu_k}<r_k
 <4\log\frac{\mu_{k+1}}{\mu_k}.
 \label{eq:app-exact-clock-r-log-comparison}
\end{equation}
Together with \eqref{eq:app-exact-clock-log-increment-two-sided} and
\eqref{eq:app-exact-clock-r-log-comparison}, this gives
\begin{equation}
 \frac{(1+a)\log K}{K(1+\omega_k)}
 <r_k<
 \frac{8(1+a)\log K}{K(1+\omega_k)}.
 \label{eq:app-exact-clock-r-order}
\end{equation}
Substituting this estimate into
\(\beta_k=4/(3r_{k-1})\) gives
\begin{equation}
 \frac{K(1+\omega_{k-1})}{6(1+a)\log K}
 <\beta_k<
 \frac{4K(1+\omega_{k-1})}{3(1+a)\log K},
 \qquad 2\leq k\leq K.
 \label{eq:app-exact-clock-beta-order}
\end{equation}

\eqref{eq:exact-smooth-omega} and \eqref{eq:app-exact-clock-ratio-bound} give
\begin{equation}
 \frac{\log2}{1+\log2}(1+a\log K)
 \leq\max\{a\log K,\log2\}
 <\omega_0,
 \qquad
 \frac{\omega_0}{2}<\omega_1<\omega_0\leq1+a\log K,
 \label{eq:app-exact-clock-initial-comparison}
\end{equation}
% At the final endpoint, \eqref{eq:exact-smooth-omega} and
% \eqref{eq:app-exact-clock-ratio-bound} give
and
\begin{equation}
 \omega_K=\log(1+K^{-1}),
 \qquad
 \omega_K<\omega_{K-1}<2\omega_K.
 \label{eq:app-exact-clock-terminal-comparison}
\end{equation}
For \(t\geq0\),
\[
 \frac{t}{1+t}
 \leq\int_0^t\frac{ds}{1+s}
 =\log(1+t)
 \leq t.
\]
Taking \(t=K^{-1}\) gives
\begin{equation}
 \frac1{2K}\leq\log(1+K^{-1})\leq\frac1K.
 \label{eq:app-exact-log-one-plus-inverse}
\end{equation}
\eqref{eq:app-exact-clock-initial-comparison},
\eqref{eq:app-exact-clock-terminal-comparison}, and
\eqref{eq:app-exact-log-one-plus-inverse} imply
\begin{equation}
 \omega_0=\Theta(\omega_1)=\Theta(1+a\log K),
 \qquad
 \omega_{K-1}=\Theta(\omega_K)=\Theta(K^{-1}),
 \label{eq:app-exact-clock-endpoint-orders}
\end{equation}
where all implicit constants are absolute.  Substitution into
\eqref{eq:app-exact-clock-beta-order} gives
\begin{equation}
  \beta_2=\Theta\!\left(
  \frac{K}{1+a}\left(a+\frac1{\log K}\right)\right),\qquad
  \beta_K=\Theta\!\left(\frac{K}{(1+a)\log K}\right).
 \label{eq:app-exact-beta-endpoint-orders}
\end{equation}
Together with \eqref{eq:app-exact-beta-monotonicity}, this proves
part~\textup{(b)} and \eqref{eq:exact-smooth-beta-endpoint-orders}.

\paragraph{Step 3: unimodality and peak location of \(\lambda_k\).}
To prove the shape and location claims for \((\lambda_k)_{k=1}^K\), we express the
sequence as samples of a one-variable profile at the decreasing grid points
\((\omega_k)_{k=1}^K\).
For \(w>0\), define
\begin{equation}
 \Delta(w)=\log\!\left(
 K^{(a+1)/K}
 -\bigl(K^{(a+1)/K}-1\bigr)e^{-w}
 \right),
 \qquad
 \Phi(w)=\frac{3\Delta(w)^2}{w+2\Delta(w)}.
 \label{eq:app-exact-lambda-profile-functions}
\end{equation}
For \(1\leq k\leq K\), \eqref{eq:exact-smooth-omega} gives
\(e^{\omega_k}-1=K^{a-(a+1)k/K}\), and hence
\begin{equation}
 \omega_{k-1}
 =\log\!\left(1+K^{(a+1)/K}(e^{\omega_k}-1)\right)
 =\omega_k+\Delta(\omega_k),
 \qquad
 r_k=\frac{2\Delta(\omega_k)}{\omega_k}.
 \label{eq:app-exact-lambda-profile-clock}
\end{equation}
\eqref{eq:app-exact-lambda-profile-clock},
\eqref{eq:exact-smooth-r}, and
\eqref{eq:exact-smooth-ikm-parameters} give
\begin{equation}
 \lambda_k
 =\frac{K}{4(1+1/a)\omega_0}\,\Phi(\omega_k).
 \label{eq:app-exact-lambda-profile-identity}
\end{equation}

We next show that \(\Phi\) has a unique maximizer.  Differentiating the
second formula in \eqref{eq:app-exact-lambda-profile-functions} gives
\begin{equation}
 \Phi'(w)
 =\frac{3\Delta(w)}{(w+2\Delta(w))^2}
 \left[2(w+\Delta(w))\Delta'(w)-\Delta(w)\right].
 \label{eq:app-exact-lambda-profile-derivative}
\end{equation}
The prefactor in \eqref{eq:app-exact-lambda-profile-derivative} is positive
for \(w>0\), so only the bracket determines the sign.  By \eqref{eq:app-exact-lambda-profile-functions} we have
\begin{equation}
 \begin{aligned}
  \Delta'(w)
  &=\frac{\bigl(K^{(a+1)/K}-1\bigr)e^{-w}}
  {K^{(a+1)/K}-\bigl(K^{(a+1)/K}-1\bigr)e^{-w}}>0,
  \qquad \Delta''(w)=-\Delta'(w)(1+\Delta'(w))<0.
 \end{aligned}
 \label{eq:app-exact-lambda-delta-derivatives}
\end{equation}
Using \eqref{eq:app-exact-lambda-delta-derivatives}, the derivative of the
bracket in \eqref{eq:app-exact-lambda-profile-derivative} is
\begin{equation}
 \begin{aligned}
 &\frac{d}{dw}
 \left[2(w+\Delta(w))\Delta'(w)-\Delta(w)\right]=\Delta'(w)(1+\Delta'(w))
 \left[
 2(1-w-\Delta(w))-\frac1{1+\Delta'(w)}
 \right].
 \end{aligned}
 \label{eq:app-exact-lambda-bracket-derivative}
\end{equation}
The bracket on the RHS of
\eqref{eq:app-exact-lambda-bracket-derivative} is strictly decreasing, since
\begin{equation}
 \begin{aligned}
 \frac{d}{dw}
 \left[
 2(1-w-\Delta(w))-\frac1{1+\Delta'(w)}
 \right]&=-2(1+\Delta'(w))
 +\frac{\Delta''(w)}{(1+\Delta'(w))^2}\\
 &\overset{\eqref{eq:app-exact-lambda-delta-derivatives}}=-2(1+\Delta'(w))
 -\frac{\Delta'(w)}{1+\Delta'(w)}<0.
 \end{aligned}
 \label{eq:app-exact-lambda-derivative-sign-control}
\end{equation}
The bracket in \eqref{eq:app-exact-lambda-profile-derivative} and the
sign-controlling bracket in
\eqref{eq:app-exact-lambda-bracket-derivative} have the following values at
\(w=0\) and limits as \(w\to\infty\):
\begin{equation}
 \begin{aligned}
 \left[2(w+\Delta(w))\Delta'(w)-\Delta(w)\right]_{w=0}
 &=0,\\
 \left[
 2(1-w-\Delta(w))-\frac1{1+\Delta'(w)}
 \right]_{w=0}
 &=2-K^{-(a+1)/K}>0,\\
 \lim_{w\to\infty}
 \left[2(w+\Delta(w))\Delta'(w)-\Delta(w)\right]
 &=-\frac{(a+1)\log K}{K}<0,\\
 \lim_{w\to\infty}
 \left[
 2(1-w-\Delta(w))-\frac1{1+\Delta'(w)}
 \right]
 &=-\infty.
 \end{aligned}
 \label{eq:app-exact-lambda-bracket-endpoints}
\end{equation}
\eqref{eq:app-exact-lambda-bracket-derivative}--
\eqref{eq:app-exact-lambda-bracket-endpoints} show that the bracket in
\eqref{eq:app-exact-lambda-profile-derivative} first increases from zero
and then decreases to a negative limit, crossing zero exactly once on
\((0,\infty)\).  Thus \(\Phi'\) is positive and then negative, so
\(\Phi\) has a unique maximizer \(w_\star\in(0,\infty)\).

By \eqref{eq:app-exact-lambda-profile-identity}, \(\lambda_k\) is a positive
constant times \(\Phi(\omega_k)\), and
\eqref{eq:app-exact-clock-first-derivative} shows that \(\omega_k\) decreases
strictly with \(k\).  Hence the sampled sequence increases until the grid
crosses \(w_\star\) and decreases afterward.  If \(w_\star\) lies outside the sampled
interval \([\omega_K,\omega_1]\), the maximum is attained at one boundary index; otherwise it is
attained at one or both of the adjacent grid points bracketing \(w_\star\).
This proves the asserted unimodality for every \(K\geq2\) and \(a\)
satisfying \eqref{eq:exact-smooth-a-range}.

It remains to locate \(w_\star\).  By
\eqref{eq:exact-smooth-a-range},
\begin{equation}
 0<\frac{(a+1)\log K}{K}\leq\log2.
 \label{eq:app-exact-lambda-step-uniform}
\end{equation}
The first identity in \eqref{eq:app-exact-lambda-delta-derivatives} also
admits the form
\begin{equation}
 \Delta'(w)=\frac{1-e^{-\Delta(w)}}{e^w-1}
 =\frac{\Delta(w)}{e^w-1}\phi(\Delta(w)),
 \label{eq:app-exact-lambda-delta-phi}
\end{equation}
where \(\phi\) is defined in \eqref{eq:app-exact-clock-phi}.  Hence the
bracket in \eqref{eq:app-exact-lambda-profile-derivative} equals
\begin{equation}
 \Delta(w)\left[
 \frac{2(w+\Delta(w))}{e^w-1}\phi(\Delta(w))-1
 \right].
 \label{eq:app-exact-lambda-bracket-phi}
\end{equation}
By \eqref{eq:app-exact-clock-phi} and
\eqref{eq:app-exact-clock-phi-comparison},
\(1/(1+s)\leq\phi(s)<1\) for \(s>0\).  Thus
\eqref{eq:app-exact-lambda-bracket-phi} is positive at \(w=1\):
\begin{equation}
 \frac{2(1+\Delta(1))}{e-1}\phi(\Delta(1))-1
 \geq\frac2{e-1}-1>0.
 \label{eq:app-exact-lambda-bracket-at-one}
\end{equation}
Moreover, \eqref{eq:app-exact-lambda-profile-functions} and
\eqref{eq:app-exact-lambda-step-uniform} give
\(0<\Delta(2)<(a+1)\log K/K\leq\log2<1\), so the same bracket is negative
at \(w=2\):
\begin{equation}
 \frac{2(2+\Delta(2))}{e^2-1}\phi(\Delta(2))-1
 <\frac6{e^2-1}-1<0.
 \label{eq:app-exact-lambda-bracket-at-two}
\end{equation}
The uniqueness established in
\eqref{eq:app-exact-lambda-bracket-derivative}--
\eqref{eq:app-exact-lambda-bracket-endpoints} therefore gives
\begin{equation}
 1<w_\star<2.
 \label{eq:app-exact-lambda-peak-window}
\end{equation}

Taylor's theorem and \eqref{eq:app-exact-lambda-step-uniform} give, with an
absolute remainder constant,
\begin{equation}
 K^{(a+1)/K}-1
 =\exp\!\left(\frac{(a+1)\log K}{K}\right)-1
 =\frac{(a+1)\log K}{K}
 +O\!\left(\frac{(a+1)^2\log^2 K}{K^2}\right).
 \label{eq:app-exact-lambda-clock-factor-expansion}
\end{equation}
Moreover, \eqref{eq:app-exact-lambda-profile-functions} and the first
identity in \eqref{eq:app-exact-lambda-delta-derivatives} can be written as
\begin{equation}
 \begin{aligned}
  \Delta(w)
  &=\log\!\left(
  1+\bigl(K^{(a+1)/K}-1\bigr)(1-e^{-w})
  \right),\\
  \Delta'(w)
  &=\frac{\bigl(K^{(a+1)/K}-1\bigr)e^{-w}}
  {1+\bigl(K^{(a+1)/K}-1\bigr)(1-e^{-w})}.
 \end{aligned}
 \label{eq:app-exact-lambda-expansion-forms}
\end{equation}
For \(w\in[1,2]\),
\eqref{eq:app-exact-lambda-clock-factor-expansion} and
\((1+x)^{-1}=1+O(x)\) give
\begin{equation}
 \frac{1}
 {1+\bigl(K^{(a+1)/K}-1\bigr)(1-e^{-w})}
 =1+O\!\left(\frac{(a+1)\log K}{K}\right).
 \label{eq:app-exact-lambda-reciprocal-expansion}
\end{equation}
Applying \(\log(1+x)=x+O(x^2)\) to the first identity in
\eqref{eq:app-exact-lambda-expansion-forms}, and using
\eqref{eq:app-exact-lambda-clock-factor-expansion} and
\eqref{eq:app-exact-lambda-reciprocal-expansion} in the second, gives
\begin{equation}
 \begin{aligned}
  \Delta(w)
  &=\frac{(a+1)\log K}{K}(1-e^{-w})
  +O\!\left(\frac{(a+1)^2\log^2 K}{K^2}\right),\\
  \Delta'(w)
  &=\left[
  \frac{(a+1)\log K}{K}
  +O\!\left(\frac{(a+1)^2\log^2 K}{K^2}\right)
  \right]e^{-w}\cdot
  \left[
  1+O\!\left(\frac{(a+1)\log K}{K}\right)
  \right]\\
  &=\frac{(a+1)\log K}{K}e^{-w}
  +O\!\left(\frac{(a+1)^2\log^2 K}{K^2}\right).
 \end{aligned}
 \label{eq:app-exact-lambda-delta-expansion}
\end{equation}
Substituting \eqref{eq:app-exact-lambda-delta-expansion} into the bracket in
\eqref{eq:app-exact-lambda-profile-derivative} yields, uniformly for
\(w\in[1,2]\),
\begin{equation}
 \begin{aligned}
  \frac{K}{(a+1)\log K}
  \left[2(w+\Delta(w))\Delta'(w)-\Delta(w)\right]
  &=q(w)+O\!\left(\frac{(a+1)\log K}{K}\right),\\
  q(w)&=e^{-w}(2w+1)-1.
 \end{aligned}
 \label{eq:app-exact-lambda-limiting-derivative}
\end{equation}
Then we have
\begin{equation}
 \begin{gathered}
  q'(w)=e^{-w}(1-2w),\qquad q(0)=0,\\
  q(1)=3e^{-1}-1>0,\qquad q(2)=5e^{-2}-1<0.
 \end{gathered}
 \label{eq:app-exact-lambda-limiting-signs}
\end{equation}
Thus \(q\) increases on \((0,1/2)\), decreases on \((1/2,\infty)\), and has a
unique positive zero \(\rho_\lambda\in(1,2)\).  It satisfies
\begin{equation}
 e^{\rho_\lambda}=1+2\rho_\lambda.
 \label{eq:app-exact-lambda-peak-root}
\end{equation}
The bracket in \eqref{eq:app-exact-lambda-profile-derivative} vanishes at
\(w_\star\).  Thus \eqref{eq:app-exact-lambda-peak-window} and
\eqref{eq:app-exact-lambda-limiting-derivative} give
\begin{align}\label{eq:q'w}
 q(w_\star)=O\!\left(\frac{(a+1)\log K}{K}\right).
\end{align}
Moreover, \eqref{eq:app-exact-lambda-limiting-signs} gives the uniform bound
\begin{equation}
 |q'(w)|=e^{-w}(2w-1)\geq e^{-2},
 \qquad w\in[1,2].
 \label{eq:app-exact-lambda-q-derivative-gap}
\end{equation}
Since \(q(\rho_\lambda)=0\), \(q'\) has constant sign on \([1,2]\), and
\(w_\star,\rho_\lambda\in(1,2)\),
\[
 |q(w_\star)|
 =\left|\int_{\rho_\lambda}^{w_\star}q'(s)\,ds\right|
 \overset{\eqref{eq:app-exact-lambda-q-derivative-gap}}\geq
 e^{-2}|w_\star-\rho_\lambda|.
\]
Consequently,
\[
 |w_\star-\rho_\lambda|
 \leq e^2|q(w_\star)|
 \overset{\eqref{eq:q'w}}=O\!\left(\frac{(a+1)\log K}{K}\right).
\]
Therefore,
\begin{equation}
 w_\star
 =\rho_\lambda
 +O\!\left(\frac{(a+1)\log K}{K}\right).
 \label{eq:app-exact-lambda-continuous-peak}
\end{equation}

On the other hand, \eqref{eq:exact-smooth-omega} is equivalently
\begin{equation}
 \log(e^{\omega_k}-1)
 =\left(a-\frac{(a+1)k}{K}\right)\log K,
 \qquad 0\leq k\leq K.
 \label{eq:app-exact-lambda-linearized-clock}
\end{equation}
Thus the strictly increasing transformation
\(\omega\mapsto\log(e^\omega-1)\) maps
\((\omega_k)_{k=0}^K\) to a decreasing arithmetic grid with spacing
\((a+1)\log K/K\).  Moreover,
\[
 \frac{d}{dw}\log(e^w-1)
 =\frac1{1-e^{-w}}\leq\frac1{1-e^{-1}},
 \qquad w\in[1,2].
\]
Thus, by
\eqref{eq:app-exact-lambda-continuous-peak},
\begin{equation}
 \log(e^{w_\star}-1)
 =\log(e^{\rho_\lambda}-1)
 +O\!\left(\frac{(a+1)\log K}{K}\right)
 \overset{\eqref{eq:app-exact-lambda-peak-root}}=\log(2\rho_\lambda)
 +O\!\left(\frac{(a+1)\log K}{K}\right).
 \label{eq:app-exact-lambda-argument-peak}
\end{equation}
The left-hand side of \eqref{eq:app-exact-lambda-argument-peak} is positive
by \eqref{eq:app-exact-lambda-peak-window}, whereas
\eqref{eq:app-exact-lambda-linearized-clock} at \(k=K\) equals
\(-\log K<0\).  Replacing \(\omega_k\) by \(w_\star\) in
\eqref{eq:app-exact-lambda-linearized-clock} and solving for the real-valued
grid index gives
\[
 \frac{aK}{a+1}
 -\frac{K\log(e^{w_\star}-1)}{(a+1)\log K},
\]
which is strictly less than \(K\).  The sequence \((\lambda_k)_{k=1}^K\)
samples only the integer grid indices \(1,\ldots,K\).  If the real-valued
index above lies in \([1,K)\), then it either is an integer or lies between
two adjacent integers.  Hence
\eqref{eq:app-exact-lambda-profile-identity} and the sign change in
\eqref{eq:app-exact-lambda-profile-derivative} imply that \(k_\lambda\) is
that integer or one of those two adjacent integers.  If the real-valued
index is less than \(1\), then
\(w_\star>\omega_1\), so
\eqref{eq:app-exact-lambda-profile-identity} and
\eqref{eq:app-exact-lambda-profile-derivative} imply \(k_\lambda=1\).
Consequently,
\begin{equation}
 \left|
 k_\lambda
 -\max\!\left\{
 1,\frac{aK}{a+1}
 -\frac{K\log(e^{w_\star}-1)}{(a+1)\log K}
 \right\}
 \right|\leq1.
 \label{eq:app-exact-lambda-grid-peak}
\end{equation}
Combining \eqref{eq:app-exact-lambda-argument-peak} and
\eqref{eq:app-exact-lambda-grid-peak} yields
\begin{equation}
 k_\lambda
 =\max\!\left\{1,
 \frac{aK}{a+1}
 -\frac{\log(2\rho_\lambda)}{a+1}\frac{K}{\log K}\right\}+O(1).
 \label{eq:app-exact-lambda-index-peak}
\end{equation}
So \eqref{eq:exact-smooth-lambda-peak-location} follows.

\paragraph{Step 4: endpoint values and peak height of \(\lambda_k\).}
It remains to prove the three orders in
\eqref{eq:exact-smooth-lambda-endpoint-orders}.  First, \eqref{eq:exact-smooth-ikm-parameters} gives
\begin{equation}
 \lambda_k
 =\frac{K\omega_{k-1}}
 {4(1+1/a)\omega_0}\,u_kr_k.
 \label{eq:app-exact-lambda-mass-identity}
\end{equation}
To estimate the factor \(u_kr_k\), using \(0<r_k<2\) from
\eqref{eq:r_k_in_(0,2)} in the definition of \(u_k\) in
\eqref{eq:exact-smooth-r} gives
\begin{equation}
 \frac{r_k}{8}<u_k<\frac{3r_k}{4}.
 \label{eq:app-exact-u-local-order}
\end{equation}
\eqref{eq:app-exact-clock-r-order} and
\eqref{eq:app-exact-u-local-order} give
\begin{equation}
 \frac{(1+a)^2\log^2 K}
 {8K^2(1+\omega_k)^2}
 <u_kr_k<
 \frac{48(1+a)^2\log^2 K}
 {K^2(1+\omega_k)^2}.
 \label{eq:app-exact-ur-order}
\end{equation}
\eqref{eq:app-exact-clock-ratio-bound} gives
\begin{equation}
 \omega_{k-1}=\Theta(\omega_k).
 \label{eq:app-exact-consecutive-clock-order}
\end{equation}
The first comparison in \eqref{eq:app-exact-clock-initial-comparison} gives
\begin{equation}
 \frac{a}{(1+a)(1+a\log K)}
 \leq\frac1{(1+1/a)\omega_0}
 <\frac{1+\log2}{\log2}
 \frac{a}{(1+a)(1+a\log K)}.
 \label{eq:app-exact-lambda-normalization-order}
\end{equation}
Substituting \eqref{eq:app-exact-ur-order}--
\eqref{eq:app-exact-lambda-normalization-order} into
\eqref{eq:app-exact-lambda-mass-identity} yields
\begin{equation}
 \lambda_k
 =\Theta\!\left(
 \frac{a(1+a)\log^2 K}
 {K(1+a\log K)}
 \frac{\omega_k}{(1+\omega_k)^2}
 \right),
 \qquad 1\leq k\leq K.
 \label{eq:app-exact-clock-lambda-order}
\end{equation}
\eqref{eq:app-exact-clock-endpoint-orders} also gives
\begin{equation}
 \frac{\omega_1}{(1+\omega_1)^2}
 =\Theta\!\left(\frac1{1+a\log K}\right),
 \qquad
 \frac{\omega_K}{(1+\omega_K)^2}
 =\Theta\!\left(\frac1K\right).
 \label{eq:app-exact-lambda-endpoint-profile-orders}
\end{equation}
\eqref{eq:app-exact-clock-lambda-order} and
\eqref{eq:app-exact-lambda-endpoint-profile-orders} yield
\begin{equation}
 \begin{aligned}
  \lambda_1
  &=\Theta\!\left(
  \frac{a(1+a)\log^2 K}
  {K(1+a\log K)^2}
  \right),
  \qquad
  \lambda_K=\Theta\!\left(\frac{1+a\log K}{K}\right)\lambda_1.
 \end{aligned}
 \label{eq:app-exact-lambda-endpoint-orders}
\end{equation}

For the peak height, define
\begin{equation}
 j=\min\{1\leq k\leq K:\omega_k\leq\log2\}.
 \label{eq:app-exact-lambda-peak-index}
\end{equation}
This index exists because \eqref{eq:exact-smooth-omega} gives
\(\omega_0>\log2>\omega_K\).  By
\eqref{eq:app-exact-clock-ratio-bound} and the minimality in
\eqref{eq:app-exact-lambda-peak-index},
\begin{equation}
 \frac{\log2}{2}<\omega_j\leq\log2,
 \qquad
 \frac{\omega_j}{(1+\omega_j)^2}
 \geq\frac{\log2}{2(1+\log2)^2}.
 \label{eq:app-exact-lambda-peak-clock-window}
\end{equation}
Together with \(s/(1+s)^2\leq1/4\) for \(s>0\),
\eqref{eq:app-exact-lambda-peak-clock-window} gives
\begin{equation}
 \max_{1\leq k\leq K}
 \frac{\omega_k}{(1+\omega_k)^2}
 =\Theta(1).
 \label{eq:app-exact-lambda-profile-maximum-order}
\end{equation}
\eqref{eq:app-exact-clock-lambda-order},
\eqref{eq:app-exact-lambda-profile-maximum-order}, and
\eqref{eq:app-exact-lambda-endpoint-orders} give
\begin{equation}
 \max_{1\leq k\leq K}\lambda_k
 =\Theta(1+a\log K)\,\lambda_1.
 \label{eq:app-exact-lambda-maximum-order}
\end{equation}
\eqref{eq:app-exact-lambda-endpoint-orders} and
\eqref{eq:app-exact-lambda-maximum-order} prove
\eqref{eq:exact-smooth-lambda-endpoint-orders} and complete the proof of
Lemma~\ref{lem:exact-smooth-schedule-properties}.

\subsection{Proof of Theorem~\ref{thm:onestate-terminal-rate}}
\label{app:proof_stochastic_rate}
\paragraph{Step 1: reducing the last-iterate residual to an accumulated potential.}
For \(\rho\in\mathbb R\), define the final affine point \(v_K^\rho\),
and let \(\mathcal R_K\) be the worst-case root-mean-square residual over
the segment \(0\leq\rho\leq1\):
\begin{equation}
 v_K^\rho=(1-\rho)x_K+\rho z_K,
 \qquad
 \mathcal R_K=
 \sup_{0\leq\rho\leq1}
 \left(\E\norm{v_K^\rho-Tv_K^\rho}^2\right)^{1/2}.
 \label{eq:app-exact-terminal-segment-target}
\end{equation}
By \eqref{eq:exact-smooth-parameter-ranges}, \(\beta_K>0\).  Hence the
definitions of \(y_K\) and \(z_K\) in \eqref{eq:sikm} give
\begin{equation}
 y_K-x_K
 =\alpha_K(x_K-x_{K-1})
 =\frac{\alpha_K}{\beta_K}(z_K-x_K).
 \label{eq:app-exact-terminal-segment-identity}
\end{equation}
% Combining \eqref{eq:app-exact-terminal-segment-target} and
% \eqref{eq:app-exact-terminal-segment-identity} gives
Then, by the iKM update~\eqref{eq:sikm},
\begin{equation}
 x_K=v_K^0,
 \qquad
 y_K=v_K^{\alpha_K/\beta_K},
 \qquad
 z_K=v_K^1.
 \label{eq:app-exact-three-terminal-points}
\end{equation}
\eqref{eq:app-exact-three-terminal-points} and
\eqref{eq:app-exact-alpha-ratio} show that all three final
points $x_K,y_K,z_K$ correspond to parameters in \([0,1]\).  Therefore
\begin{equation}
 \max_{v_K\in\{x_K,y_K,z_K\}}
 \left(\E\norm{v_K-Tv_K}^2\right)^{1/2}
 \leq\mathcal R_K.
 \label{eq:app-exact-terminal-max-reduction}
\end{equation}
Thus it suffices to bound \(\mathcal R_K\).

Lemma~\ref{lem:app-residual-operator} shows that the operator $G$ in
\eqref{eq:app-residual-operator} satisfies the hypotheses of
Lemma~\ref{lem:existence}.
For every \(1\leq k\leq K\),
\eqref{eq:app-exact-proof-masses} gives \(\mu_k>0\).  Let \(p_k\) be the
unique solution of
\begin{equation}
 Gp_k+\mu_k(p_k-x_k)=0.
 \label{eq:app-exact-regularized-root}
\end{equation}
Existence and uniqueness follow from Lemma~\ref{lem:existence}.  Moreover,
\eqref{eq:regularized-root-firm-nonexpansiveness} and Cauchy--Schwarz imply
that, for every \(\mu>0\),
\[
 \norm{p_\mu(v)-p_\mu(v')}\leq\norm{v-v'}.
\]
Thus \(p_\mu\) is continuous.  Since \(x_k\) is
\(\gF_k\)-measurable, it follows that
\(p_k=p_{\mu_k}(x_k)\) is \(\gF_k\)-measurable.

By the definition of \(v_K^\rho\) in
\eqref{eq:app-exact-terminal-segment-target} and the root equation
\eqref{eq:app-exact-regularized-root},
\begin{align}
 \mu_K(v_K^\rho-p_K)
 &=(1-\rho)\mu_K(x_K-p_K)
 +\rho\mu_K(z_K-p_K)\notag\\
 &=(1-\rho)Gp_K+\rho\mu_K(z_K-p_K).
 \label{eq:app-exact-terminal-root-identity}
\end{align}
Consequently, for every \(0\leq\rho\leq1\),
\begin{align}
 \mu_K\norm{v_K^\rho-p_K}
 &\leq(1-\rho)\norm{Gp_K}
 +\rho\mu_K\norm{z_K-p_K}\notag\\
 &\leq
 \sqrt{(1-\rho)^2+\rho^2}\,
 \sqrt{\norm{Gp_K}^2+\mu_K^2\norm{z_K-p_K}^2}\notag\\
 &\leq\sqrt{\norm{Gp_K}^2+\mu_K^2\norm{z_K-p_K}^2},
 \label{eq:app-exact-terminal-distance}
\end{align}
where the second inequality is Cauchy--Schwarz in \(\mathbb R^2\), and the
third inequality uses \((1-\rho)^2+\rho^2\leq1\) for
\(0\leq\rho\leq1\).

Define the potential
\begin{equation}
 \Psi_k\coloneqq\mu_k^2\norm{z_k-p_k}^2+\norm{Gp_k}^2.
 \label{eq:app-exact-potential}
\end{equation}
Then \eqref{eq:app-exact-terminal-distance} gives
\begin{align}
  \norm{v_K^\rho-p_K}\leq \frac{1}{\mu_K}\sqrt{\Psi_K}.
\end{align}
The triangle inequality and the \(2\)-Lipschitz estimate in
\eqref{eq:app-residual-operator-properties} give
\begin{align}
 \norm{Gv_K^\rho}
 &\leq\norm{Gv_K^\rho-Gp_K}+\norm{Gp_K}\notag\\
 &\overset{\eqref{eq:app-residual-operator-properties}}\leq
 2\norm{v_K^\rho-p_K}+\norm{Gp_K}\notag\\
 &\leq\left(1+\frac{2}{\mu_K}\right)\sqrt{\Psi_K},
 \label{eq:app-exact-terminal-residual-conversion}
\end{align}
where the third inequality uses \eqref{eq:app-exact-terminal-distance} and the
immediate consequence \(\norm{Gp_K}\leq\sqrt{\Psi_K}\) of
\eqref{eq:app-exact-potential}.

For real-valued random variables, write
\[
 L^2(\Omega;\mathbb R)\coloneqq\left\{X:\Omega\to\mathbb R:
 X\text{ is measurable and }\E|X|^2<\infty\right\}.
\]
The following lemma, whose proof is given in
Appendix~\ref{app:proof-exact-potential-integrability}, ensures that the potential is integrable.
\begin{lemma}[Potential measurability and integrability]
\label{lem:app-exact-potential-integrability}
For every \(1\leq k\leq K\), the random variable \(\Psi_k\) is
\(\gF_k\)-measurable and integrable.  Consequently,
\(\sqrt{\Psi_k}\in L^2(\Omega;\mathbb R)\) and
\((\E\Psi_k)^{1/2}<\infty\).
\end{lemma}

By Lemma~\ref{lem:app-exact-potential-integrability}, taking the
\(L^2\)-norm for each fixed \(0\leq\rho\leq1\) is justified and yields
\begin{align*}
 \left(\E\norm{Gv_K^\rho}^2\right)^{1/2}
 &\leq
 \left(1+\frac{2}{\mu_K}\right)(\E\Psi_K)^{1/2}.
\end{align*}
The right-hand side is independent of \(\rho\).  Taking the supremum over
\(0\leq\rho\leq1\) and using the definition of \(\mathcal R_K\) in
\eqref{eq:app-exact-terminal-segment-target}, we obtain
\begin{equation}
 \mathcal R_K
 \leq\left(1+\frac{2}{\mu_K}\right)(\E\Psi_K)^{1/2}.
 \label{eq:app-exact-terminal-potential-reduction}
\end{equation}

\paragraph{Step 3: one-step estimate of the potential.} It therefore remains to control \((\E\Psi_K)^{1/2}\).  For this purpose,
we use the following one-step estimate.
\begin{lemma}[one-step potential estimate]
\label{lem:app-exact-one-step-transport}
For every \(1\leq k\leq K-1\),
\begin{equation}
 \E[\Psi_{k+1}\mid\gF_k]
 \leq
 \left(\sqrt{\Psi_k}+r_k\norm{b_k}\right)^2+r_k^2\sigma^2
 \qquad\text{almost surely}.
 \label{eq:app-exact-conditional-transport}
\end{equation}
\end{lemma}

Its proof appears in Appendix~\ref{app:proof-exact-one-step-transport}.  Taking
expectations in
\eqref{eq:app-exact-conditional-transport}, for
\(1\leq k\leq K-1\) we obtain
\begin{align}
 \E\Psi_{k+1}
 =\E\!\left[\E[\Psi_{k+1}\mid\gF_k]\right]&\leq
 \E\!\left[
 \left(\sqrt{\Psi_k}+r_k\norm{b_k}\right)^2
 \right]+r_k^2\sigma^2\notag\\
%  &=
%  \left\|\sqrt{\Psi_k}+r_k\norm{b_k}\right\|_{L^2(\Omega)}^2
%  +r_k^2\sigma^2\notag\\
 &\leq
%  \left(
%  \left\|\sqrt{\Psi_k}\right\|_{L^2(\Omega)}
%  +r_k\left\|\norm{b_k}\right\|_{L^2(\Omega)}
%  \right)^2+r_k^2\sigma^2\notag\\
%  &=
 \left(
 (\E\Psi_k)^{1/2}
 +r_k(\E\norm{b_k}^2)^{1/2}
 \right)^2+r_k^2\sigma^2,
 \label{eq:app-exact-mean-square-recursion}
\end{align}
where the second inequality is
Minkowski's inequality in \(L^2(\Omega)\), using \(r_k>0\) from
\eqref{eq:r_k_in_(0,2)}.  All quantities are finite by
Lemma~\ref{lem:app-exact-potential-integrability} and the finiteness of
\(B_K\) in \eqref{eq:biased-BK}.  Since both sides of
\eqref{eq:app-exact-mean-square-recursion} are nonnegative, taking square
roots gives
\begin{equation}
 (\E\Psi_{k+1})^{1/2}
 \leq
 \left\|
 \left(
 (\E\Psi_k)^{1/2}
 +r_k(\E\norm{b_k}^2)^{1/2},
 r_k\sigma
 \right)
 \right\|_2.
 \label{eq:app-exact-rms-recursion}
\end{equation}
For \(0\leq j\leq K-1\), put
\begin{equation}
 S_j=\sum_{k=1}^j r_k(\E\norm{b_k}^2)^{1/2},
 \qquad
 V_j=\sigma\left(\sum_{k=1}^j r_k^2\right)^{1/2},
 \qquad S_0=V_0=0.
 \label{eq:app-exact-accumulation-terms}
\end{equation}
We claim that
\begin{equation}
 (\E\Psi_{j+1})^{1/2}
 \leq(\E\Psi_1)^{1/2}+S_j+V_j,
 \qquad 0\leq j\leq K-1.
 \label{eq:app-exact-accumulation-induction}
\end{equation}
For \(j=0\), the claim is an equality because \(S_0=V_0=0\).
Now fix \(1\leq j\leq K-1\) and suppose that the claim holds with
\(j-1\) in place of \(j\).  The induction hypothesis and the definition
of \(S_j\) give
\begin{align}
 (\E\Psi_j)^{1/2}
 +r_j(\E\norm{b_j}^2)^{1/2}
 &\leq
 (\E\Psi_1)^{1/2}+S_{j-1}+V_{j-1}
 +r_j(\E\norm{b_j}^2)^{1/2}\notag\\
 &=(\E\Psi_1)^{1/2}+S_j+V_{j-1}.
 \label{eq:app-exact-accumulation-induction-step}
\end{align}
All quantities in
\eqref{eq:app-exact-accumulation-induction-step} are nonnegative by
\eqref{eq:app-exact-accumulation-terms} and
\eqref{eq:r_k_in_(0,2)}.
Therefore the Euclidean norm is nondecreasing in its first coordinate,
and \eqref{eq:app-exact-rms-recursion}, \eqref{eq:app-exact-accumulation-induction-step} give
\begin{align*}
 (\E\Psi_{j+1})^{1/2}
 &\leq
 \left\|
 \left(
 (\E\Psi_j)^{1/2}
 +r_j(\E\norm{b_j}^2)^{1/2},
 r_j\sigma
 \right)
 \right\|_2\\
 &\leq
 \left\|
 \left(
 (\E\Psi_1)^{1/2}+S_j+V_{j-1},
 r_j\sigma
 \right)
 \right\|_2\\
 &=
 \left\|
 \left((\E\Psi_1)^{1/2}+S_j,0\right)
 +\left(V_{j-1},r_j\sigma\right)
 \right\|_2\\
 &\leq
 \left\|\left((\E\Psi_1)^{1/2}+S_j,0\right)\right\|_2
 +\left\|\left(V_{j-1},r_j\sigma\right)\right\|_2\\
 &=(\E\Psi_1)^{1/2}+S_j+V_j,
\end{align*}
where the last inequality is the Euclidean triangle inequality, and the last
equality uses
\(V_j^2=V_{j-1}^2+r_j^2\sigma^2\), which follows from
\eqref{eq:app-exact-accumulation-terms}.
This completes the induction.  Taking \(j=K-1\) yields
\begin{equation}
 (\E\Psi_K)^{1/2}
 \leq(\E\Psi_1)^{1/2}
 +\sum_{k=1}^{K-1}r_k(\E\norm{b_k}^2)^{1/2}
 +\sigma\left(\sum_{k=1}^{K-1}r_k^2\right)^{1/2}.
 \label{eq:app-exact-accumulation}
\end{equation}

Combining \eqref{eq:app-exact-terminal-potential-reduction} and
\eqref{eq:app-exact-accumulation} yields the decomposition
\begin{equation}
 \mathcal R_K
 \leq\left(1+\frac{2}{\mu_K}\right)
 \left\{
 (\E\Psi_1)^{1/2}
 +\sum_{k=1}^{K-1}r_k(\E\norm{b_k}^2)^{1/2}
 +\sigma\left(\sum_{k=1}^{K-1}r_k^2\right)^{1/2}
 \right\}.
 \label{eq:app-exact-terminal-decomposition}
\end{equation}

\paragraph{Step 3: bounding the terms in
\eqref{eq:app-exact-terminal-decomposition}.}
We first bound \((\E\Psi_1)^{1/2}\) in
\eqref{eq:app-exact-terminal-decomposition}.
Set
\begin{equation}
 D\coloneqq\inf_{p^\star\in\Fix T}\norm{x_1-p^\star},
 \qquad 0\leq D<\infty,
 \label{eq:app-exact-initial-fixed-point-distance}
\end{equation}
where finiteness follows from \(\Fix T\ne\varnothing\).
Since \(\beta_1=0\), \eqref{eq:sikm} gives \(z_1=x_1\).  Thus substituting
\eqref{eq:app-exact-regularized-root} into
\eqref{eq:app-exact-potential} gives
\begin{equation}
 \Psi_1=2\mu_1^2\norm{x_1-p_1}^2.
 \label{eq:app-exact-initial-potential-identity}
\end{equation}
Moreover, \(x_1\) is deterministic, so uniqueness in
Lemma~\ref{lem:existence} implies that \(p_1\), and hence \(\Psi_1\),
is deterministic.  Fix any \(p^\star\in\Fix T\). 
\eqref{eq:app-residual-zero-fixed-point} and
\eqref{eq:app-exact-regularized-root} give
\(Gp^\star=0\) and \(Gp_1=\mu_1(x_1-p_1)\), respectively.  Hence the monotonicity
estimate in \eqref{eq:app-residual-operator-properties} gives
\begin{align*}
 0
 &\leq\ip{Gp_1-Gp^\star}{p_1-p^\star}
 =\mu_1\ip{x_1-p_1}{p_1-p^\star}\\
 &=\mu_1\ip{x_1-p_1}
 {(x_1-p^\star)-(x_1-p_1)}\\
 &=\mu_1\left(
\ip{x_1-p_1}{x_1-p^\star}-\norm{x_1-p_1}^2
 \right).
\end{align*}
Since \(\mu_1>0\), rearranging this inequality gives the first
inequality below, and Cauchy--Schwarz gives the second:
\begin{equation}
 \norm{x_1-p_1}^2
 \leq\ip{x_1-p_1}{x_1-p^\star}
 \leq\norm{x_1-p_1}\norm{x_1-p^\star}.
 \label{eq:app-exact-initial-cauchy-step}
\end{equation}
If \(x_1=p_1\), the desired distance bound is immediate; otherwise,
dividing \eqref{eq:app-exact-initial-cauchy-step} by
\(\norm{x_1-p_1}\) gives the same bound.  Since \(p^\star\in\Fix T\) was
arbitrary, \eqref{eq:app-exact-initial-fixed-point-distance} gives
\begin{align}
 \norm{x_1-p_1}&\leq
 \inf_{p^\star\in\Fix T}\norm{x_1-p^\star}=D,\notag\\
 (\E\Psi_1)^{1/2}
 &\overset{\eqref{eq:app-exact-initial-potential-identity}}=
 \sqrt2\,\mu_1\norm{x_1-p_1}\overset{\eqref{eq:app-exact-proof-masses}}=
 \frac{4\sqrt2(1+1/a)}{K}\norm{x_1-p_1}
 \leq\frac{4\sqrt2(1+1/a)D}{K}.
 \label{eq:app-exact-initial-potential-bound}
\end{align}

We next bound
\(\sum_{k=1}^{K-1}r_k(\E\norm{b_k}^2)^{1/2}\) and
\(\sigma(\sum_{k=1}^{K-1}r_k^2)^{1/2}\) in \eqref{eq:app-exact-terminal-decomposition}.
Squaring \eqref{eq:app-exact-clock-r-uniform} and summing over
\(k=1,\ldots,K\) yields
\begin{align}
 \sum_{k=1}^K r_k^2
 &\leq
 K\left(
 \frac{2(a+1)K^{(a+1)/K}\log K}{K}
 \right)^2=\frac{4(a+1)^2K^{2(a+1)/K}\log^2K}{K}.
 \label{eq:app-exact-clock-r-square-sum-bound}
\end{align}
\eqref{eq:app-exact-clock-r-uniform} and the definition of
\(B_K\) in \eqref{eq:biased-BK} give
\begin{align}
 \sum_{k=1}^{K-1}r_k(\E\norm{b_k}^2)^{1/2}
 &\leq
 \left(\max_{1\leq k\leq K-1}r_k\right)
 \sum_{k=1}^{K-1}(\E\norm{b_k}^2)^{1/2}\leq
 \frac{2(a+1)K^{(a+1)/K}\log K}{K}
 B_K.
 \label{eq:app-exact-bias-bound}
\end{align}
\eqref{eq:app-exact-clock-r-square-sum-bound} gives
\begin{equation}
 \sigma\left(\sum_{k=1}^{K-1}r_k^2\right)^{1/2}
 \leq
 \frac{2(a+1)K^{(a+1)/K}\sigma\log K}{\sqrt K}.
 \label{eq:app-exact-noise-bound}
\end{equation}
At iteration \(K\), \eqref{eq:exact-smooth-omega} gives
\[
 (1+1/a)\omega_0>(a+1)\log K,
 \qquad
 \omega_K=\log(1+K^{-1})<K^{-1},
\]Therefore
\begin{align}
 0
 &\overset{\eqref{eq:app-exact-proof-masses}}<\frac2{\mu_K}
 \overset{\eqref{eq:app-exact-proof-masses}}=
 \frac{K\omega_{K-1}}{2(1+1/a)\omega_0}\overset{\eqref{eq:app-exact-clock-ratio-bound}}<
 \frac{K\omega_K}{(1+1/a)\omega_0}
<
 \frac{K\omega_K}{(a+1)\log K}<
 \frac1{(a+1)\log K}<\frac1{\log2},
 \label{eq:app-exact-terminal-mass-bound}
\end{align}
where the last inequality uses \(a>0\) and \(K\geq2\).
Substituting \eqref{eq:app-exact-initial-potential-bound},
\eqref{eq:app-exact-bias-bound}, \eqref{eq:app-exact-noise-bound}, and
\eqref{eq:app-exact-terminal-mass-bound} into
\eqref{eq:app-exact-terminal-decomposition} gives
\begin{align}
 \mathcal R_K
 &\leq
 \left(1+\frac1{\log2}\right)
 \left\{
 \frac{4\sqrt2(1+1/a)D}{K}
 +2(a+1)K^{(a+1)/K}\left(
 \frac{\sigma\log K}{\sqrt K}
 +\frac{B_K\log K}{K}
 \right)
 \right\}\notag\\
 &\leq\left(1+\frac1{\log2}\right)
 \left\{
 \frac{4\sqrt2(1+1/a)D}{K}
 +4(a+1)\left(
 \frac{\sigma\log K}{\sqrt K}
 +\frac{B_K\log K}{K}
 \right)
 \right\},
 \label{eq:app-exact-before-constants}
\end{align}
where the last inequality uses \(K^{(a+1)/K}\leq2\) from \eqref{eq:app-exact-clock-ratio-bound}.
% The two numerical coefficients satisfy
% \begin{equation}
%  \frac{39}{32}\,4\sqrt2=\frac{39\sqrt2}{8}<7,
%  \qquad
%  \frac{39}{32}\,4=\frac{39}{8}<5.
%  \label{eq:app-exact-final-coefficient-bounds}
% \end{equation}
% Combining \eqref{eq:app-exact-before-constants},
% \eqref{eq:app-exact-final-elementary-bounds}, and
% \eqref{eq:app-exact-final-coefficient-bounds} gives
% \begin{align}
%  \mathcal R_K
%  &\leq
%  7\left(1+\frac1a\right)\frac{D}{K}
% +5(1+a)\left(
%  \frac{\sigma\log K}{\sqrt K}
%  +\frac{B_K\log K}{K}
%  \right).%\notag\\
% %  &\leq7\left\{
% %  \left(1+\frac1a\right)\frac{D}{K}+(1+a)\left(
% %  \frac{\sigma\log K}{\sqrt K}
% %  +\frac{B_K\log K}{K}
% %  \right)\right\}.
%  \label{eq:app-exact-final-segment-bound}
% \end{align}
% % Finally, for each \(v_K\in\{x_K,y_K,z_K\}\),
% % \begin{equation}
% %  \norm{Gv_K}
% %  \overset{\eqref{eq:app-residual-operator}}=\norm{v_K-Tv_K}.
% %  \label{eq:app-exact-residual-conversion}
% % \end{equation}
\eqref{eq:app-exact-initial-fixed-point-distance},
\eqref{eq:app-exact-terminal-max-reduction} and
\eqref{eq:app-exact-before-constants} give
\eqref{eq:onestate-terminal-rate}.

\subsection{Proofs of the auxiliary lemmas}
\label{app:proof-exact-smooth-auxiliary-lemmas}

\subsubsection{Proof of Lemma~\ref{lem:app-residual-operator}}
\label{app:proof-residual-operator}
By
\eqref{eq:app-residual-operator}, it gives
\begin{align}
 &2\ip{Gx-Gy}{x-y}-\norm{Gx-Gy}^2\notag\\
 &=2\ip{x-y-(Tx-Ty)}{x-y}
 -\norm{x-y-(Tx-Ty)}^2\notag\\
 &=\norm{x-y}^2-\norm{Tx-Ty}^2
 \overset{\eqref{eq:stochastic-nonexpansiveness}}\geq0.
 \label{eq:app-residual-cocoercivity-identity}
\end{align}
This is the first inequality in
\eqref{eq:app-residual-operator-properties}.  Cauchy--Schwarz then gives
\begin{equation}
 \frac12\norm{Gx-Gy}^2
 \leq\ip{Gx-Gy}{x-y}
 \leq\norm{Gx-Gy}\norm{x-y},
 \label{eq:app-residual-lipschitz-derivation}
\end{equation}
which yields the second inequality in
\eqref{eq:app-residual-operator-properties}.  Finally,
\begin{equation}
 Gp^\star=0
 \overset{\eqref{eq:app-residual-operator}}\Longleftrightarrow
 p^\star-Tp^\star=0
 \Longleftrightarrow p^\star\in\Fix T,
\end{equation}
which proves \eqref{eq:app-residual-zero-fixed-point}.
% The Lipschitz constant is sharp, since
% \begin{equation}
%  T=-I
%  \quad\Longrightarrow\quad
%  G=2I,
%  \qquad
%  \norm{Gx-Gy}=2\norm{x-y}.
% \end{equation}

\subsubsection{Proof of Lemma~\ref{lem:existence}}
\label{app:proof-existence}
Since \(G\) is continuous and monotone with full domain, it is maximal
monotone~\citep[Corollary~20.28]{bauschke2017convex}.  Minty's theorem therefore shows that its resolvent is everywhere
defined, and
\begin{equation}
 p_\mu(v)\coloneqq (I+\mu^{-1}G)^{-1}v
 \label{eq:app-regularized-root-resolvent}
\end{equation}
exists uniquely for every \(v\in\gH\).  If \(p=p_\mu(v)\) and
\(p'=p_\mu(v')\), their defining equations give
\begin{equation}
 Gp=\mu(v-p),\qquad Gp'=\mu(v'-p'),
\end{equation}
and therefore
\begin{equation}\label{eq:Gp=Gp'}
 Gp-Gp'=\mu\bigl\{(v-v')-(p-p')\bigr\}.
\end{equation}
Further, monotonicity of \(G\) gives
\begin{equation}
 0\leq\ip{Gp-Gp'}{p-p'}
 \overset{\eqref{eq:Gp=Gp'}}=\mu\left\{\ip{v-v'}{p-p'}-\norm{p-p'}^2\right\}.
\end{equation}
Therefore
\[
 \norm{p_\mu(v)-p_\mu(v')}^2
 \leq\ip{p_\mu(v)-p_\mu(v')}{v-v'},
\]
which is \eqref{eq:regularized-root-firm-nonexpansiveness}.

\subsubsection{Proof of Lemma~\ref{lem:app-exact-schedule-bounds}}
\label{app:proof-exact-schedule-bounds}

For \(x>0\) and \(c>1\), the function
\(x\mapsto(1+x)^c-1-cx\) vanishes at zero and has derivative
\(c((1+x)^{c-1}-1)>0\).  Therefore
\begin{equation}
 \log(1+cx)<c\log(1+x),\qquad \forall x>0,\,\,c>1.
 \label{eq:app-exact-log-scaling}
\end{equation}
For \(x=K^{a-(a+1)k/K}\) and \(c=K^{(a+1)/K}\), we have
\(x>0\), \(c>1\), and \eqref{eq:exact-smooth-omega} gives
\[
 \omega_k=\log(1+x),
 \qquad
 \omega_{k-1}=\log(1+cx).
\]
Thus by
\eqref{eq:app-exact-log-scaling},
\(\omega_{k-1}<c\omega_k\) which yields
\[
 1<\frac{\omega_{k-1}}{\omega_k}<K^{(a+1)/K}.
\]
This together with \eqref{eq:exact-smooth-a-range} proves
\eqref{eq:app-exact-clock-ratio-bound}.  Taking logarithms and using the definition of \(\mu_k\) in \eqref{eq:app-exact-proof-masses}, we have
\begin{equation}
 0<\log\frac{\mu_{k+1}}{\mu_k}
 <\frac{(a+1)\log K}{K}.
 \label{eq:app-exact-clock-log-ratio-bound}
\end{equation}
\eqref{eq:app-exact-schedule-conversion} and
\eqref{eq:app-exact-clock-ratio-bound} together give
\eqref{eq:r_k_in_(0,2)}.

For \(q>1\), the function \(q\log q-(q-1)\) vanishes at one and has
derivative \(\log q>0\).  Thus
\eqref{eq:app-exact-schedule-conversion},
\eqref{eq:app-exact-clock-ratio-bound}, and
\eqref{eq:app-exact-clock-log-ratio-bound} give
\begin{align*}
 r_k
 &=2\left(\frac{\mu_{k+1}}{\mu_k}-1\right)\leq2\frac{\mu_{k+1}}{\mu_k}
 \log\frac{\mu_{k+1}}{\mu_k}
 <\frac{2(a+1)K^{(a+1)/K}\log K}{K},
\end{align*}
which proves \eqref{eq:app-exact-clock-r-uniform}.

To prove \eqref{eq:app-exact-clock-step-ratio-uniform-gap}, set
\begin{equation}
 \Delta_k=\omega_{k-1}-\omega_k,
 \qquad
 \zeta_k=K^{a-(a+1)(k-1)/K}.
 \label{eq:Delta,zeta}
\end{equation}
Then \eqref{eq:exact-smooth-omega} gives
\begin{align}
 \Delta_k
 &=\log(1+\zeta_k)
 -\log\!\left(1+\zeta_ke^{-(a+1)\log K/K}\right)\notag\\
 &=-\int_0^{(a+1)\log K/K}
 \frac{d}{ds}\log(1+\zeta_ke^{-s})\,ds\notag\\
 &=\int_0^{(a+1)\log K/K}
 \frac{\zeta_ke^{-s}}{1+\zeta_ke^{-s}}\,ds\notag\\
 &<\frac{(a+1)\log K}{K}\frac{\zeta_k}{1+\zeta_k}.
 \label{eq:app-exact-clock-delta-local-bound}
\end{align}
Also, \eqref{eq:exact-smooth-omega} gives
\begin{equation}
  (1+1/a)\omega_0>(1+1/a)a\log K=(a+1)\log K.
  \label{eq:app-exact-clock-mass-lower-bound}
 \end{equation}
Since \(r_k=2\Delta_k/\omega_k\) by \eqref{eq:exact-smooth-r},
\eqref{eq:app-exact-proof-masses},
\eqref{eq:app-exact-clock-delta-local-bound}, and
\eqref{eq:app-exact-clock-mass-lower-bound} yield
\begin{align}
 \frac{r_k}{\mu_k}
 &=\frac{K\omega_{k-1}\Delta_k}
 {2(1+1/a)\omega_0\omega_k}<\frac12\frac{\omega_{k-1}}{\omega_k}
 \frac{\zeta_k}{1+\zeta_k}.
 \label{eq:app-exact-clock-step-ratio-local}
\end{align}
If \(0<\zeta_k\leq4\), then
\eqref{eq:app-exact-clock-ratio-bound} and
\eqref{eq:app-exact-clock-step-ratio-local} give
\begin{equation}
 \frac{r_k}{\mu_k}<\frac12\cdot2\cdot\frac45=\frac45.
 \label{eq:app-exact-clock-step-ratio-small-zeta}
\end{equation}
If \(\zeta_k>4\), then
\eqref{eq:exact-smooth-a-range}, \eqref{eq:exact-smooth-omega}
and \eqref{eq:app-exact-clock-delta-local-bound}
 give
\begin{equation}
 \Delta_k
 <\log2\frac{\zeta_k}{1+\zeta_k}<\log2,
 \qquad
 \omega_k
 =\log\!\left(1+\frac{\zeta_k}{K^{(a+1)/K}}\right)>\log3.
 \label{eq:app-exact-clock-large-zeta-bounds}
\end{equation}
By \eqref{eq:app-exact-clock-large-zeta-bounds},
\begin{equation}
 \frac{\omega_{k-1}}{\omega_k}
 =1+\frac{\Delta_k}{\omega_k}
 <1+\frac{\log2}{\log3}<\frac53.
 \label{eq:app-exact-clock-large-zeta-ratio}
\end{equation}
\eqref{eq:app-exact-clock-step-ratio-local} and
\eqref{eq:app-exact-clock-large-zeta-ratio} imply
\[
 \frac{r_k}{\mu_k}
 <\frac12\cdot\frac53\frac{\zeta_k}{1+\zeta_k}<\frac56.
\]
Together with
\eqref{eq:app-exact-clock-step-ratio-small-zeta} and the positivity in
\eqref{eq:app-exact-proof-masses} and \eqref{eq:r_k_in_(0,2)}, this proves
\eqref{eq:app-exact-clock-step-ratio-uniform-gap}.

By \eqref{eq:r_k_in_(0,2)} and \eqref{eq:exact-smooth-r}, we have
\begin{equation}
  0<u_k=
  \frac{3r_k}{2(1+r_k)(2+r_k)}<
  \frac{3r_k}{2(2+r_k)}<\frac34<1,
  \label{eq:app-exact-u-range}
 \end{equation} 
\eqref{eq:app-exact-u-range},
\eqref{eq:app-exact-schedule-conversion}, and
\eqref{eq:app-exact-clock-step-ratio-uniform-gap} give
\eqref{eq:app-exact-lambda-range}.  

For \(2\leq k\leq K\),
\eqref{eq:exact-smooth-ikm-parameters} and
\eqref{eq:r_k_in_(0,2)} give \(\beta_k>0\), while
\eqref{eq:exact-smooth-ikm-parameters} and
\eqref{eq:app-exact-schedule-conversion} give
\[
 \frac{\alpha_k}{\beta_k}
 =\frac{u_k-\lambda_k}{1-\lambda_k}
 =u_k\frac{\mu_k-r_k}{\mu_k-u_kr_k}.
\]
By \eqref{eq:app-exact-lambda-range}, this ratio is positive, and
\[
 u_k-\frac{\alpha_k}{\beta_k}
 =\frac{\lambda_k(1-u_k)}{1-\lambda_k}>0.
\]
The above two expressions prove \eqref{eq:app-exact-alpha-ratio}.

\subsubsection{Proof of
Lemma~\ref{lem:app-exact-potential-integrability}}
\label{app:proof-exact-potential-integrability}

For \(1\leq k\leq K-1\), the conditional second-moment bound in
\eqref{eq:biased-centered-conditions} and the tower property for the
nonnegative random variable \(\norm{\xi_k}^2\) give
\begin{equation}
 \E\norm{\xi_k}^2
 =\E\!\left[\E[\norm{\xi_k}^2\mid\gF_k]\right]
 \leq\sigma^2.
 \label{eq:app-exact-noise-L2}
\end{equation}
Thus \(\xi_k\in L^2\).  Moreover, the finiteness of \(B_K\) in
\eqref{eq:biased-BK} gives \(b_k\in L^2\) for the same range of \(k\).

The triangle inequality and \eqref{eq:stochastic-nonexpansiveness} give
\begin{equation}
 \norm{Tz}
 \leq\norm{Tz-T0}+\norm{T0}
 \overset{\eqref{eq:stochastic-nonexpansiveness}}\leq
 \norm z+\norm{T0}.
\end{equation}
Consequently,
\begin{equation}
 \E\norm{Tz}^2
 \leq2\E\norm z^2+2\norm{T0}^2<\infty
 \qquad\text{whenever }z\in L^2.
 \label{eq:app-exact-map-L2}
\end{equation}
Thus \(T\) maps \(L^2\) random variables to \(L^2\) random variables.

The query formulas in \eqref{eq:sikm} can be written, for
\(1\leq k\leq K\), as
\begin{equation}
 \begin{aligned}
 y_k&=(1+\alpha_k)x_k-\alpha_kx_{k-1},\\
 z_k&=(1+\beta_k)x_k-\beta_kx_{k-1}.
 \end{aligned}
 \label{eq:app-exact-L2-query-formulas}
\end{equation}
For \(1\leq k\leq K-1\),
\eqref{eq:biased-T-oracle} and \eqref{eq:sikm} further give
\begin{equation}
 \begin{aligned}
 \widehat T_k(z_k)&=Tz_k+b_k+\xi_k,\\
 x_{k+1}&=(1-\lambda_k)y_k+\lambda_k\widehat T_k(z_k).
 \end{aligned}
 \label{eq:app-exact-L2-update-formulas}
\end{equation}
Since \(L^2\) is closed under finite linear combinations and all schedule
coefficients are deterministic and finite by
\eqref{eq:exact-smooth-omega}--\eqref{eq:exact-smooth-ikm-parameters} and
\eqref{eq:exact-smooth-parameter-ranges},
\eqref{eq:app-exact-map-L2}--\eqref{eq:app-exact-L2-update-formulas}
show that
\begin{equation}
 x_{k-1},x_k,b_k,\xi_k\in L^2
 \quad\Longrightarrow\quad
 y_k,z_k,Tz_k,\widehat T_k(z_k),x_{k+1}\in L^2.
 \label{eq:app-exact-L2-propagation}
\end{equation}
Since the arbitrary initial points \(x_0,x_1\) are deterministic, they lie in
\(L^2\).  Induction over \(k=1,\ldots,K-1\) therefore proves
\(x_0,\ldots,x_K\in L^2\).  The two query formulas in
\eqref{eq:app-exact-L2-query-formulas} then imply \(y_k,z_k\in L^2\) for
every \(1\leq k\leq K\).  Notice that this
argument uses neither the \(K\)-th oracle call nor any
square-integrability assumption on \(b_K\).

Fix \(1\leq k\leq K\), and let \(p_{\mu_k}(\cdot)\) denote the
regularized-root map from Lemma~\ref{lem:existence}.  Equation
\eqref{eq:app-exact-regularized-root} gives
\(p_k=p_{\mu_k}(x_k)\).  Firm nonexpansiveness
\eqref{eq:regularized-root-firm-nonexpansiveness} and Cauchy--Schwarz give
\begin{align*}
 \norm{p_k-p_{\mu_k}(0)}^2
 &\leq\ip{p_k-p_{\mu_k}(0)}{x_k}
 \leq\norm{p_k-p_{\mu_k}(0)}\norm{x_k}.
\end{align*}
Since both norms are nonnegative, it follows pointwise that
\begin{equation}
 \norm{p_k-p_{\mu_k}(0)}\leq\norm{x_k}
 \qquad\text{almost surely}.
 \label{eq:app-exact-root-L2-control}
\end{equation}
Since \(x_k\in L^2\) and \(p_{\mu_k}(0)\in\gH\) is deterministic,
\eqref{eq:app-exact-root-L2-control} implies \(p_k\in L^2\).
Hence \(z_k-p_k\in L^2\), while
\eqref{eq:app-exact-regularized-root} implies \(Gp_k\in L^2\).
It follows from \eqref{eq:app-exact-potential} that
\begin{equation}
 \E\Psi_k
 =\mu_k^2\E\norm{z_k-p_k}^2+\E\norm{Gp_k}^2<\infty.
 \label{eq:app-exact-potential-integrability}
\end{equation}

Finally, \(z_k\) is \(\gF_k\)-measurable by the oracle setup, and
\(p_k\) is \(\gF_k\)-measurable by the argument following
\eqref{eq:app-exact-regularized-root}.
\eqref{eq:app-residual-operator-properties} implies that \(G\) is
continuous, and hence \(Gp_k\) is
\(\gF_k\)-measurable.  Therefore \(\Psi_k\) is
\(\gF_k\)-measurable.  Equation
\eqref{eq:app-exact-potential-integrability} is equivalent to
\(\sqrt{\Psi_k}\in L^2(\Omega;\mathbb R)\), and it also gives
\((\E\Psi_k)^{1/2}<\infty\).

\subsubsection{Proof of Lemma~\ref{lem:app-exact-one-step-transport}}
\label{app:proof-exact-one-step-transport}

Define the corrected query point
\begin{equation}
 \widetilde z_k
 \coloneqq z_k-\frac{r_k}{\mu_k}
 \bigl(z_k-\widehat T_k(z_k)\bigr).
 \label{eq:app-exact-corrected-query}
\end{equation}
Fix \(1\leq k\leq K-1\).
The definitions of \(u_k\) and \(\alpha_k\) in
\eqref{eq:exact-smooth-r} and
\eqref{eq:exact-smooth-ikm-parameters}, respectively, give
\begin{equation}
 (1-\lambda_k)\alpha_k+\lambda_k\beta_k
 =(u_k-\lambda_k)\beta_k+\lambda_k\beta_k
 =u_k\beta_k.
 \label{eq:app-exact-coefficient-balance}
\end{equation}
The definition of \(z_k\) in \eqref{eq:sikm} gives
\(z_k-x_k=\beta_k(x_k-x_{k-1})\).  Using this identity, we can rewrite the
\(x\)-update in \eqref{eq:sikm} as follows:
\begin{align}
 x_{k+1}
 &=x_k+(1-\lambda_k)\alpha_k(x_k-x_{k-1})
 +\lambda_k\bigl(\widehat T_k(z_k)-x_k\bigr)\notag\\
 &=x_k+
 \bigl[(1-\lambda_k)\alpha_k+\lambda_k\beta_k\bigr]
 (x_k-x_{k-1})
 -\lambda_k\bigl(z_k-\widehat T_k(z_k)\bigr)\notag\\
 &=x_k+u_k(z_k-x_k)
 -u_k\frac{r_k}{\mu_k}
 \bigl(z_k-\widehat T_k(z_k)\bigr)\notag\\
 &=(1-u_k)x_k+u_k\widetilde z_k,
 \label{eq:app-exact-hidden-x-recursion}
\end{align}
where the second and third equalities use
\(z_k-x_k=\beta_k(x_k-x_{k-1})\), the third also uses
\eqref{eq:app-exact-coefficient-balance} and
\(\lambda_k=u_kr_k/\mu_k\) from
\eqref{eq:app-exact-schedule-conversion}, and the last equality uses
\eqref{eq:app-exact-corrected-query}.

Define the half-iterate error
\begin{equation}
 \mathcal S_k=\mu_k(\widetilde z_k-p_k)+r_kGp_k,
 \label{eq:app-exact-local-transport-notation}
\end{equation}
where $p_k$ is defined in \eqref{eq:app-exact-regularized-root}.
We'll use \(\mathcal S_k\) as a bridge to prove \eqref{eq:app-exact-conditional-transport}: we first upper bound $\Psi_{k+1}$ by $\norm{\mathcal S_k}^2+\norm{Gp_{k}}^2$, and then upper bound $\norm{\mathcal S_k}^2+\norm{Gp_{k}}^2$ in terms of $\Psi_k$.

\paragraph{Step 1: upper bounding $\Psi_{k+1}$ by $\norm{\mathcal S_k}^2+\norm{Gp_{k+1}}^2$.}
\eqref{eq:app-exact-local-transport-notation} and
\eqref{eq:app-exact-regularized-root} give
\begin{equation}
 \mu_k(\widetilde z_k-x_k)=\mathcal S_k-(1+r_k)Gp_k.
 \label{eq:app-exact-direct-displacements}
\end{equation}
Then we have
\begin{align}\label{eq:G_p_k+1}
 Gp_{k+1}&\overset{\eqref{eq:app-exact-regularized-root}}=\mu_{k+1}(x_{k+1}-p_{k+1})\notag\\
 &\overset{\eqref{eq:app-exact-hidden-x-recursion}}=
 \mu_{k+1}\left\{(1-u_k)(x_k-p_k)
 +u_k(\widetilde z_k-p_k)-(p_{k+1}-p_k)\right\}\notag\\
 &\overset{\eqref{eq:app-exact-regularized-root}}=
 \frac{\mu_{k+1}}{\mu_k}\left\{(1-u_k)Gp_k
 +u_k\mu_k(\widetilde z_k-p_k)-\mu_k(p_{k+1}-p_k)\right\}\notag\\
 &\overset{\eqref{eq:app-exact-local-transport-notation}}=
 \frac{\mu_{k+1}}{\mu_k}\left\{(1-u_k)Gp_k
 +u_k(\mathcal S_k-r_kGp_k)-\mu_k(p_{k+1}-p_k)\right\}\notag\\
 &=\frac{\mu_{k+1}}{\mu_k}
 \left\{[1-(1+r_k)u_k]Gp_k
 +u_k\mathcal S_k-\mu_k(p_{k+1}-p_k)\right\}.
\end{align}
From \eqref{eq:app-exact-schedule-conversion},
\eqref{eq:exact-smooth-r} and
\eqref{eq:exact-smooth-ikm-parameters} we can compute that
\begin{equation}
 \frac{\mu_{k+1}}{\mu_k}u_k
 =\frac{3r_k}{4(1+r_k)},\qquad
 \frac{\mu_{k+1}}{\mu_k}\beta_{k+1}u_k
 =\frac1{1+r_k},
 \qquad
 1-\frac{\mu_{k+1}}{\mu_k}
 \bigl[1-(1+r_k)u_k\bigr]=\frac{r_k}{4}.
 \label{eq:app-exact-direct-scalar-balances}
\end{equation}
Using \eqref{eq:app-exact-direct-scalar-balances} in
\eqref{eq:G_p_k+1} yields
\begin{equation}
 \mu_{k+1}(p_{k+1}-p_k)
 =\frac{3r_k}{4(1+r_k)}\mathcal S_k-\frac{r_k}{4}Gp_k
 -(Gp_{k+1}-Gp_k).
\label{eq:app-exact-direct-root-transport}
\end{equation}
Consequently,
\begin{align}
 0
 &\leq2\mu_{k+1}
 \ip{p_{k+1}-p_k}{Gp_{k+1}-Gp_k}\notag\\
 &\overset{\eqref{eq:app-exact-direct-root-transport}}=\frac{3r_k}{2(1+r_k)}
 \ip{\mathcal S_k}{Gp_{k+1}-Gp_k}
 -\frac{r_k}{2}\ip{Gp_k}{Gp_{k+1}-Gp_k}-2\norm{Gp_{k+1}-Gp_k}^2,
 \label{eq:app-exact-direct-monotone-input}
\end{align}
where the inequality follows from the monotonicity estimate in
\eqref{eq:app-residual-operator-properties}, together with
\(\mu_{k+1}>0\) from
\eqref{eq:app-exact-proof-masses}.
Since \(r_k>0\) by \eqref{eq:r_k_in_(0,2)}, solving
\eqref{eq:app-exact-direct-monotone-input} for the cross term gives
\begin{equation}
 2\ip{Gp_k}{Gp_{k+1}-Gp_k}
 \leq\frac{6}{1+r_k}\ip{\mathcal S_k}{Gp_{k+1}-Gp_k}
 -\frac8{r_k}\norm{Gp_{k+1}-Gp_k}^2.
 \label{eq:app-exact-direct-cross-term-bound}
\end{equation}
Moreover, substituting \eqref{eq:app-exact-hidden-x-recursion} into the update
\(z_{k+1}=x_{k+1}+\beta_{k+1}(x_{k+1}-x_k)\)
from \eqref{eq:sikm} gives
\begin{equation}
 z_{k+1}-x_{k+1}=\beta_{k+1}u_k(\widetilde z_k-x_k),\quad z_{k+1}=x_k+(1+\beta_{k+1})u_k(\widetilde z_k-x_k).
 \label{eq:app-exact-direct-z-recursion}
\end{equation}
Thus we have
\begin{align}
 \mu_{k+1}(z_{k+1}-p_{k+1})
 &\overset{\eqref{eq:app-exact-regularized-root}}=\mu_{k+1}(z_{k+1}-x_{k+1})+Gp_{k+1}\notag\\
 &\overset{\eqref{eq:app-exact-direct-z-recursion}}=
 \mu_{k+1}\beta_{k+1}u_k(\widetilde z_k-x_k)+Gp_{k+1}\notag\\
 &\overset{\eqref{eq:app-exact-direct-displacements}}=
 \frac{\mu_{k+1}}{\mu_k}\beta_{k+1}u_k
 \bigl[\mathcal S_k-(1+r_k)Gp_k\bigr]+Gp_{k+1}\notag\\
 &\overset{\eqref{eq:app-exact-direct-scalar-balances}}=
 \frac1{1+r_k}\bigl[\mathcal S_k-(1+r_k)Gp_k\bigr]
 +Gp_{k+1}\notag\\
 &=\frac1{1+r_k}\mathcal S_k+Gp_{k+1}-Gp_k.
 \label{eq:app-exact-direct-anchor-transport}
\end{align}
Therefore, we have the following estimation of the potential:
\begin{align}
 \Psi_{k+1}
 &=\norm{\frac1{1+r_k}\mathcal S_k+Gp_{k+1}-Gp_k}^2
 +\norm{Gp_{k+1}}^2\notag\\
 &=\frac1{(1+r_k)^2}\norm{\mathcal S_k}^2+\norm{Gp_k}^2+\frac2{1+r_k}\ip{\mathcal S_k}{Gp_{k+1}-Gp_k}\notag\\
 &\qquad+2\ip{Gp_k}{Gp_{k+1}-Gp_k}+2\norm{Gp_{k+1}-Gp_k}^2\notag\\
 &\overset{\eqref{eq:app-exact-direct-cross-term-bound}}{\leq}
 \frac1{(1+r_k)^2}\norm{\mathcal S_k}^2+\norm{Gp_k}^2+\frac8{1+r_k}\ip{\mathcal S_k}{Gp_{k+1}-Gp_k}
 -\frac{2(4-r_k)}{r_k}\norm{Gp_{k+1}-Gp_k}^2\notag\\
 &=\frac{4+7r_k}{(4-r_k)(1+r_k)^2}
 \norm{\mathcal S_k}^2+\norm{Gp_k}^2-\frac{2(4-r_k)}{r_k}
 \left\|Gp_{k+1}-Gp_k-
 \frac{2r_k}{(4-r_k)(1+r_k)}\mathcal S_k\right\|^2,
 \label{eq:app-exact-direct-square-completion}
\end{align}
where the first equality uses \eqref{eq:app-exact-direct-anchor-transport} in
the first term of the potential \eqref{eq:app-exact-potential}.
%For the second equality, write
% \(Gp_{k+1}=Gp_k+(Gp_{k+1}-Gp_k)\) and expand both squared norms.  The
% inequality then applies \eqref{eq:app-exact-direct-cross-term-bound}
% exactly to the term
% \(2\ip{Gp_k}{Gp_{k+1}-Gp_k}\).  Thus the coefficient of
% \(\ip{\mathcal S_k}{Gp_{k+1}-Gp_k}\) becomes
% \[
%  \frac2{1+r_k}+\frac6{1+r_k}=\frac8{1+r_k},
% \]
% while the coefficient of \(\norm{Gp_{k+1}-Gp_k}^2\) becomes
% \[
%  2-\frac8{r_k}=-\frac{2(4-r_k)}{r_k}.
% \]
% Finally, the last equality completes the square in
% \(Gp_{k+1}-Gp_k\).  Explicitly,
% \begin{align*}
%  &\frac8{1+r_k}\ip{\mathcal S_k}{Gp_{k+1}-Gp_k}
%  -\frac{2(4-r_k)}{r_k}\norm{Gp_{k+1}-Gp_k}^2\\
%  &\quad=
%  -\frac{2(4-r_k)}{r_k}
%  \left\|Gp_{k+1}-Gp_k-
%  \frac{2r_k}{(4-r_k)(1+r_k)}\mathcal S_k\right\|^2\\
%  &\qquad
%  +\frac{8r_k}{(4-r_k)(1+r_k)^2}\norm{\mathcal S_k}^2.
% \end{align*}
% Combining this identity with
% \[
%  \frac1{(1+r_k)^2}
%  +\frac{8r_k}{(4-r_k)(1+r_k)^2}
%  =\frac{4+7r_k}{(4-r_k)(1+r_k)^2}
% \]
% gives the last equality in
% \eqref{eq:app-exact-direct-square-completion}.
By \eqref{eq:r_k_in_(0,2)},
\((4-r_k)(1+r_k)^2>0\), and the coefficient of
\(\norm{\mathcal S_k}^2\) is at most one because
\begin{equation}
 (4-r_k)(1+r_k)^2-(4+7r_k)=r_k^2(2-r_k)\geq0.
 \label{eq:app-exact-direct-factor-domination}
\end{equation}
Thus \eqref{eq:app-exact-direct-square-completion} and
\eqref{eq:app-exact-direct-factor-domination}, together with
\(2(4-r_k)/r_k>0\), imply
\begin{equation}
 \Psi_{k+1}\leq\norm{\mathcal S_k}^2+\norm{Gp_k}^2.
 \label{eq:app-exact-direct-transport-bound}
\end{equation}

\paragraph{Step 2: upper bounding $\norm{\mathcal S_k}^2+\norm{Gp_{k+1}}^2$ in terms of $\Psi_k$.}
\eqref{eq:app-exact-local-transport-notation},
\eqref{eq:app-exact-corrected-query}, \eqref{eq:biased-T-oracle}, and
\eqref{eq:app-residual-operator} give the exact response identity
\begin{align}
 \mathcal S_k
 &\overset{\eqref{eq:app-exact-local-transport-notation}}=
 \mu_k(\widetilde z_k-p_k)+r_kGp_k\notag\\
 &\overset{\eqref{eq:app-exact-corrected-query}}=
 \mu_k(z_k-p_k)-r_k\bigl(z_k-\widehat T_k(z_k)\bigr)+r_kGp_k\notag\\
 &\overset{\eqref{eq:biased-T-oracle}}=
 \mu_k(z_k-p_k)-r_k(z_k-Tz_k-b_k-\xi_k)+r_kGp_k\notag\\
 &\overset{\eqref{eq:app-residual-operator}}=
 \mu_k(z_k-p_k)-r_k(Gz_k-Gp_k)+r_kb_k+r_k\xi_k.
 \label{eq:app-exact-response-identity}
\end{align}
Moreover, \eqref{eq:app-residual-operator} gives
\begin{equation}
 I-\frac{r_k}{\mu_k}G
 =\left(1-\frac{r_k}{\mu_k}\right)I+\frac{r_k}{\mu_k}T.
 \label{eq:app-exact-averaged-map}
\end{equation}
By \eqref{eq:app-exact-clock-step-ratio-uniform-gap}, the two coefficients on
the right are nonnegative and sum to one.  Hence, for any \(x,y\in\gH\),
\begin{align}
 &\norm{\left(I-\frac{r_k}{\mu_k}G\right)x
 -\left(I-\frac{r_k}{\mu_k}G\right)y}\notag\\
 &\quad\overset{\eqref{eq:app-exact-averaged-map}}=
 \norm{\left(1-\frac{r_k}{\mu_k}\right)(x-y)
 +\frac{r_k}{\mu_k}(Tx-Ty)}\notag\\
 &\quad\leq
 \left(1-\frac{r_k}{\mu_k}\right)\norm{x-y}
 +\frac{r_k}{\mu_k}\norm{Tx-Ty}
 \overset{\eqref{eq:stochastic-nonexpansiveness}}{\leq}\norm{x-y}.
\end{align}
Thus \(I-(r_k/\mu_k)G\) is nonexpansive.  Since \(\mu_k>0\),
\begin{align}
 &\norm{\mu_k(z_k-p_k)-r_k(Gz_k-Gp_k)}\notag\\
 &\quad=\mu_k\norm{\left(I-\frac{r_k}{\mu_k}G\right)z_k
 -\left(I-\frac{r_k}{\mu_k}G\right)p_k}
 \leq\mu_k\norm{z_k-p_k}.
 \label{eq:app-exact-deterministic-response-contraction}
\end{align}

From the measurability argument in the proof of
Lemma~\ref{lem:app-exact-potential-integrability} and the oracle setup,
the vector
\[
 \mu_k(z_k-p_k)-r_k(Gz_k-Gp_k)+r_kb_k
\]
is \(\gF_k\)-measurable.  By the same lemma,
\eqref{eq:app-exact-potential},
\eqref{eq:app-exact-deterministic-response-contraction}, and
the finiteness of \(B_K\) in \eqref{eq:biased-BK}, this vector is also
square-integrable; moreover, \(\xi_k\in L^2\) by
\eqref{eq:app-exact-noise-L2}.  Thus
\eqref{eq:app-exact-response-identity},
\eqref{eq:biased-centered-conditions}, and
\eqref{eq:app-exact-deterministic-response-contraction} imply
\begin{align}
 \E\left[\norm{\mathcal S_k}^2\middle|\gF_k\right]&=
 \norm{\mu_k(z_k-p_k)-r_k(Gz_k-Gp_k)+r_kb_k}^2
 +r_k^2\E[\norm{\xi_k}^2\mid\gF_k]\notag\\
 &\leq
 \left(\mu_k\norm{z_k-p_k}+r_k\norm{b_k}\right)^2+r_k^2\sigma^2.
 \label{eq:app-exact-conditional-response-bound}
\end{align}
Consequently,
\begin{align}
 \E\left[\norm{\mathcal S_k}^2\middle|\gF_k\right]
 +\norm{Gp_k}^2&\overset{\eqref{eq:app-exact-conditional-response-bound}}{\leq}
 \left(\mu_k\norm{z_k-p_k}+r_k\norm{b_k}\right)^2
 +\norm{Gp_k}^2+r_k^2\sigma^2\notag\\
 &\overset{\eqref{eq:app-exact-potential}}=\Psi_k
 +2\mu_kr_k\norm{z_k-p_k}\norm{b_k}
 +r_k^2\norm{b_k}^2+r_k^2\sigma^2\notag\\
 &\overset{\eqref{eq:app-exact-potential}}\leq
 \Psi_k+2r_k\sqrt{\Psi_k}\norm{b_k}
 +r_k^2\norm{b_k}^2+r_k^2\sigma^2\notag\\
 &=\left(\sqrt{\Psi_k}+r_k\norm{b_k}\right)^2+r_k^2\sigma^2.
\end{align}
% \E[\Psi_{k+1}\mid\gF_k]
%  &\overset{\eqref{eq:app-exact-direct-transport-bound}}\leq
This together with \eqref{eq:app-exact-direct-transport-bound} gives \eqref{eq:app-exact-conditional-transport}.